\documentclass[10pt,twocolumn,letterpaper]{article}

\usepackage[pagenumbers]{cvpr} 

\usepackage[dvipsnames]{xcolor}

\newcommand{\revise}[1]{{\color{black}#1}}

\usepackage[most]{tcolorbox}
\definecolor{uiHeader}{RGB}{40, 116, 166}
\usepackage{amsmath}
\usepackage{multirow}
\usepackage{booktabs}
\usepackage{pifont}
\usepackage{makecell}
\usepackage{graphicx}
\usepackage{adjustbox}
\usepackage{wrapfig}
\usepackage{cuted}

\newcommand{\xmark}{\ding{55}}

\newcommand{\ccmark}{\textcolor{OliveGreen}{\ding{51}}}
\newcommand{\xxmark}{\textcolor{Mahogany}{\ding{55}}}

\definecolor{darkgreen}{RGB}{0, 158, 96} 

\newtcolorbox{promptbox}[1]{%
  enhanced,
  colback=white,
  colframe=black!40,
  boxrule=0.8pt,
  arc=3pt,
  width=\linewidth,
  left=6pt, right=6pt, top=4pt, bottom=4pt,
  title={#1},
  colbacktitle=uiHeader,
  coltitle=white,
  fonttitle=\bfseries\small,
  fontupper=\footnotesize,
  toptitle=4pt,
  bottomtitle=4pt,
}
\definecolor{cvprblue}{rgb}{0.21,0.49,0.74}
\usepackage[pagebackref,breaklinks,colorlinks,allcolors=cvprblue]{hyperref}

\def\puppet{\textsc{Puppeteer}}
\def\sceneges{\textsc{SceneGes}}

\title{%
  \makebox[\textwidth][c]{%
  \hspace{-1.0em}
    \raisebox{-0.3\height}{%
      \includegraphics[
        height=1.2cm,
        keepaspectratio
      ]{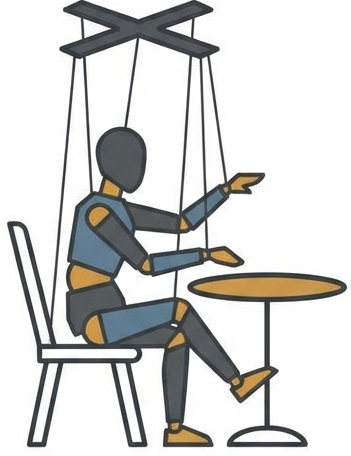}%
    }%
    \hspace{0.0em}%
    \parbox[c]{0.99\textwidth}{%
      \centering
      \textsc{Puppeteer}: Object-Grounded Posture-Aware Co-Speech Gesture Generation%
    }%
  }%
}

\author{ 
    Vida Adeli$^{1,2,3}$, 
    Soroush Mehraban$^{1,2,3}$, 
    Jacob Rommann$^{1}$, 
    Harrison Sanborn$^{1}$, \\
    Cole Clifford$^{1}$, 
    Babak Taati$^{1,2}$ \\[5pt] 
        {
        $^{1}${\textit{Pickford AI}} \; \;
        $^{2}${\textit{University of Toronto}} \; \;
        $^{3}${\textit{Vector Institute}} } \\[5pt]
        \tt \small \textbf{\href{https://puppeteer.pickford.ai/}{https://puppeteer.pickford.ai/}}
}

\begin{document}
\maketitle

\begin{strip}\centering
    \vspace*{-1.5cm}
    \captionsetup{type=figure}
    \includegraphics[width=1.0\textwidth]{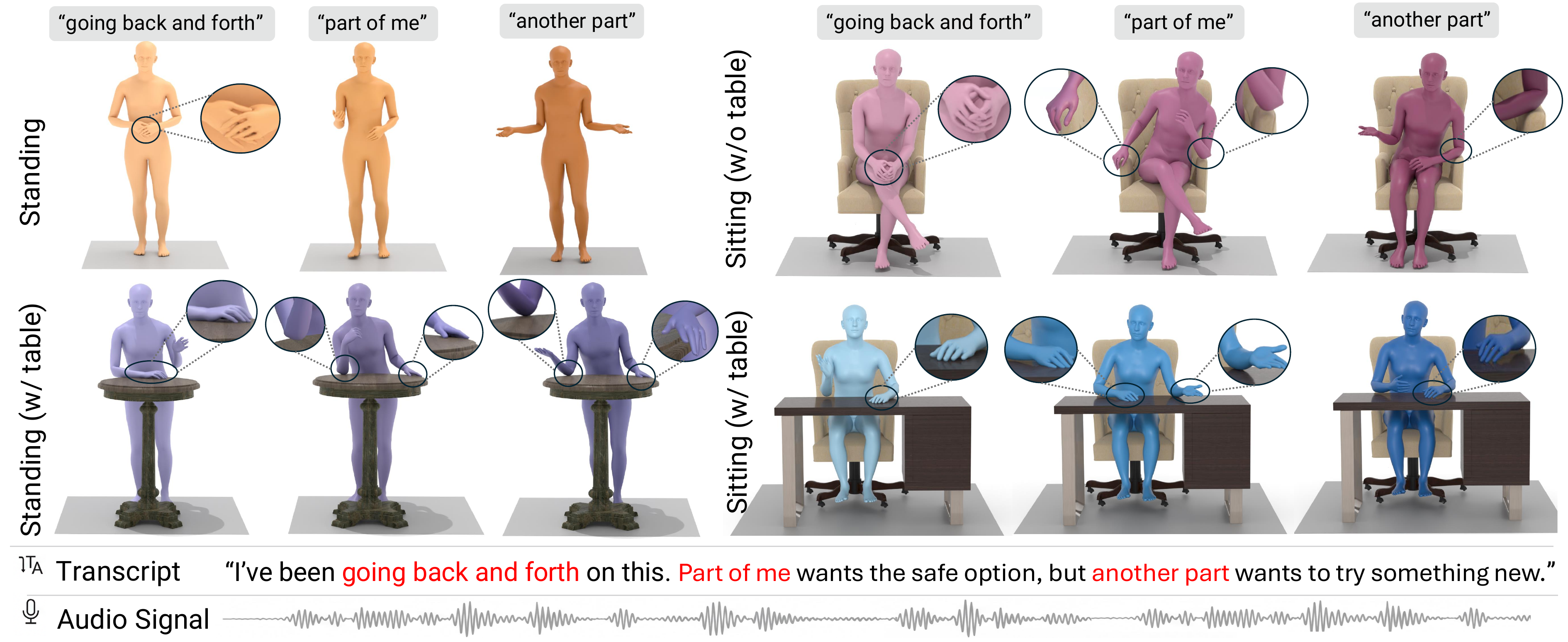}
    \vspace{-0.3in}
    \captionof{figure}{\textsc{Puppeteer} generates posture-aware and object-grounded co-speech gestures. Driven by the same audio and text transcript, the synthesized gestures semantically align with the speech while naturally adapting to different physical postures and surrounding objects.}
    \label{fig:teaser}
\end{strip}

\begin{abstract}
Generating co-speech gestures that are temporally coherent, semantically aligned with speech, and grounded with surrounding objects remains challenging. Prior speech-driven gesture models emphasize audio-gesture alignment but do not explicitly account for posture constraints or surrounding objects, failing to capture the inherent correlation between body gestures and the physical space.
We present \puppet{}, a posture-aware, object-grounded co-speech gesture diffusion model operating in a causal latent space. We decompose long gestures into structured primitives and learn a causal variational autoencoder that encodes them into temporally ordered latent tokens, each depending only on the past.
We then perform conditional diffusion directly in the causal latent space, conditioning on speech signals, motion history, an initial posture reference, and object geometry to synthesize physically consistent gestures.
This temporally ordered latent formulation enables explicit temporal control and supports tasks such as gesture in-betweening and gesture completion.
To better assess co-speech gesture synthesis beyond existing measures, we introduce new evaluation metrics tailored to this task.
We also created \sceneges, the first curated synthetic 3D dataset of embodied co-speech gestures and corresponding 3D objects, enabling object-grounded gesture generation.
Experiments show that \puppet{} generates more diverse and temporally synchronized gestures than prior methods, while enabling object-grounded gesture synthesis. 
\end{abstract}

\section{Introduction}\label{sec:intro}\vspace{-5pt}
Co-speech gestures are a crucial component of natural human communication. 
For embodied agents in everyday environments, gesture generation requires more than audio-text alignment; gestures must be temporally coherent, synchronized with speech, and physically consistent with the surrounding space (e.g., nearby furniture).

Existing speech-driven gesture models~\cite{chen2025language, liu2025semges, liu24emage, yang2025gesturehydra, chen2024diffsheg} are scene-agnostic, focusing on multimodal fusion of audio and text, and are unable to generate gestures that are spatially consistent with posture and surrounding objects.
Long-horizon generation also requires temporal coherence while maintaining precise speech alignment. 
Prior transformer-based methods often use discrete VQ-VAE tokenization, enabling token-level modeling but introducing quantization errors. Diffusion methods~\cite{yang2025gesturehydra, chen2024diffsheg, ijcai2023p650} 
operate directly in \emph{motion space} to preserve fine-grained temporal alignment but requiring denoising in a computationally demanding high-dimensional space. Latent diffusion gesture models~\cite{ao2023gesturediffuclip, chen2024Synerg} 
reduce this cost, but their bidirectional latent representations entangle temporal information, limiting explicit control over speech–gesture alignment.

In parallel, scene-aware motion generation has advanced physically grounded human–scene interaction modeling, but primarily targets locomotion or task-oriented motions rather than communicative co-speech gestures~\cite{yi2024generating, ghosh2026scemos, yi2022human, zhao2023synthesizing}. As a result, these methods do not capture the posture-dependent and communicative nature of everyday gesturing.
Recent work~\cite{rajan2025interactalker} incorporates object interaction into gesture generation through separate pretraining on gesture and interaction tasks, treating gesture formation and environmental interaction as modular components. 
Its gesture model is trained on the standing-only, scene-agnostic BEAT2 dataset~\cite{liu24emage}, limiting posture–gesture modeling, while object interaction is learned independently and integrated later, assuming gesture and environment can be composed post hoc.
However, communicative gestures are inherently shaped by both posture and nearby objects; different body configurations, even within sitting or standing, constrain gesture formation (posture-awareness), while nearby furniture, e.g., tables or armrests, further shapes spatial configuration (object-awareness).
\revise{
Crucially, we use \emph{object awareness} to refer to the influence of surrounding physical structures on conversational gestures, rather than action-centric object manipulation. Nearby furniture constrains gesture space, affects resting hand placement, and shapes posture-dependent motion. For example, a speaker may keep their hands above a table, rest an arm on an armrest, or reduce gesture amplitude when seated near furniture. Our goal is therefore not to generate manipulation actions such as grasping or lifting, but to model how passive scene geometry shapes natural co-speech gestures.}

To address posture-awareness, we leverage Embody3D~\cite{mclean2025embody}, benchmarking it for the first time for co-speech gesture generation across diverse posture conditions. 
To model object-aware gesture formation, we introduce \sceneges, enabling object-grounded gesture synthesis in structured environments.
We further introduce \puppet, a posture-aware, object-grounded co-speech gesture diffusion model operating in a causal latent primitive space.
We decompose long gesture sequences into fixed-length primitives, encode each primitive into compact continuous latent tokens with explicit temporal ordering, and perform conditional diffusion directly in this causal latent space. 
This enables autoregressive generation in causal latent space, supporting tasks such as gesture in-betweening and completion, while keeping diffusion efficient and avoiding VQ discretization artifacts.
To improve speech–gesture synchronization, we introduce lightweight temporal cross-attention control via (i) an audio-window constraint for rhythmic alignment and (ii) a text-span constraint for word-level semantic alignment.
Finally, to ground gestures in real scenes, we add an object-aware fusion module that injects geometric context while preserving the strong speech-gesture prior learned from large-scale data.
As summarized in~\cref{tab:comparison_features}, prior methods cover only subsets of these properties, while \puppet{} unifies them in a single framework.

In summary, our contributions are: 
1) The first work to address object-grounded co-speech gesture generation, modeling the intrinsic coupling between communicative gestures, posture, and surrounding objects.
2) 
A temporally controlled cross-attention masking for precise audio and word-level text alignment, enabled by adapting causal latent autoregressive diffusion to co-speech gesture generation.
3) Posture-aware training using \emph{conversational} data across diverse pose references, explicitly modeling posture–gesture coupling.
4) The \sceneges{} and new evaluation metrics for object-grounded co-speech gesture generation.


\begin{table}[t]
\centering
\scriptsize
\setlength{\tabcolsep}{1pt}
\caption{\small Comparison of co-speech gesture generation methods.} \vspace{-10pt}
\label{tab:comparison_features}
\begin{tabular}{lccccccc}
\toprule
\textbf{Method} &
\makecell{\revise{\textbf{Quantization-free }}\\\revise{\textbf{latent Space}}} &
\makecell{\textbf{Compact}\\\textbf{Space}} &
\makecell{\textbf{Causal}\\\textbf{Latent}} &
\makecell{\textbf{Diffusion}\\\textbf{Modeling}} &
\makecell{\textbf{Pose}\\\textbf{aware}} &
\makecell{\textbf{Object}\\\textbf{aware}} \\
\midrule
EMAGE~\cite{liu24emage}       & \xxmark & \ccmark & \xxmark & \xxmark  & \xxmark & \xxmark \\
SynTalker~\cite{chen2024Synerg}     & \xxmark & \ccmark & \xxmark & \ccmark & \ccmark & \xxmark \\
DiffSHEG~\cite{chen2024diffsheg}     & \ccmark & \xxmark & \xxmark & \ccmark  & \xxmark & \xxmark \\
LOM~\cite{chen2025language}         & \xxmark & \ccmark & \xxmark & \xxmark  & \ccmark & \xxmark \\
SemGes~\cite{liu2025semges}       & \xxmark & \ccmark & \xxmark & \xxmark  & \xxmark & \xxmark \\
GestureLSM~\cite{liu2025gesturelsm}  & \xxmark & \ccmark & \xxmark & \ccmark  & \xxmark & \xxmark \\
GestureHYDRA~\cite{yang2025gesturehydra} & \ccmark & \xxmark & \xxmark & \ccmark  & \xxmark & \xxmark \\
ViBES~\cite{zhang2025vibes}        & \xxmark & \ccmark & \xxmark & \xxmark  & \ccmark & \xxmark \\
EchoAvatar~\cite{chen2026echo}
& \xxmark & \ccmark & \ccmark & \xxmark & \xxmark & \xxmark \\
MIBURI~\cite{mughal2026miburi} & \xxmark & \ccmark & \ccmark & \xxmark & \xxmark & \xxmark \\
\midrule
\textbf{\puppet{} (Ours)}  & \ccmark & \ccmark & \ccmark & \ccmark & \ccmark & \ccmark \\
\bottomrule
\end{tabular}
\vspace{-10pt}
\end{table}

\section{Related Work}
\vspace{-5pt}
We present the core related work here, with an extended discussion in Appendix. 

\vspace{5pt}
\noindent \textbf{Speech-driven Gesture Generation} aims to synthesize motion aligned with speech using multimodal signals such as audio and text. Prior works explore different modelings for integrating speech and motion. Multimodal fusion approaches such as CaMN~\cite{liu22beat} progressively combine audio, text, emotion, and speaker identity signals to generate expressive gestures, while DisCo~\cite{liu2022disco} improves gesture diversity by disentangling rhythmic structure from semantic content. EMAGE~\cite{liu24emage} and The Language of Motion~\cite{chen2025language} extend gesture synthesis toward holistic multimodal modeling by jointly learning motion with other modalities or treating motion as a language modeling problem. SemGes~\cite{liu2025semges} and ViBES~\cite{zhang2025vibes} further emphasize semantic grounding and conversational context. Diffusion-based methods such as DiffuseStyleGesture~\cite{yang2023diffusestylegesture}, SynTalker~\cite{chen2024enabling}, InteracTalker~\cite{rajan2025interactalker}, ConvoFusion~\cite{mughal2024convofusion}, DiffSHEG~\cite{chen2024diffsheg}, GestureLSM~\cite{liu2025gesturelsm}, and GestureHYDRA~\cite{yang2025gesturehydra} explore stochastic denoising frameworks for gesture generation and controllability. However, most existing approaches remain scene-agnostic and do not explicitly model interactions with surrounding objects.

\vspace{5pt}
\noindent \textbf{Motion Generation.} \textbf{\textit{Scene-aware}} motion generation focuses on physically consistent human motion in 3D environments. Methods such as DIMOS~\cite{zhao2023synthesizing}, MOVER~\cite{yi2022human}, TeSMo~\cite{yi2024generating}, SceMoS~\cite{ghosh2026scemos} and HSI-GPT~\cite{wang2025hsi} model human–scene interactions through reinforcement learning, optimization, or diffusion, but primarily target locomotion or task-oriented interactions rather than communicative gestures.
\textbf{\textit{Streaming motion.}}
Recent work has explored causal and streaming motion generation. MotionStreamer~\cite{xiao2025motionstreamer} introduces a Causal VAE for text-conditioned streaming motion generation, which inspires our streaming formulation. 
However, its generative modeling and target task differ fundamentally from ours. It is scene-agnostic and text-conditioned, without audio synchronization, word-level speech alignment, posture variation, or 3D object geometry, and therefore cannot serve as a direct baseline for our task.
MIBURI~\cite{mughal2026miburi} develops an online causal framework for co-speech gestures and EchoAvatar~\cite{chen2026echo} generates continuous body motion from streaming audio. We build on this broader causal/streaming paradigm, focusing instead on posture- and object-grounded co-speech gesture generation with explicit latent-scale temporal alignment between speech dynamics and gesture primitives.

\section{\puppet{} Architecture}
\label{sec:method}

We propose a latent diffusion model for autoregressive gesture generation. The proposed model contains a causal variational autoencoder (CausalVAE) that compresses the gesture primitives into a compact latent space and a latent denoising diffusion model that predicts clean latent variables from noise, conditioned on speech signals, gesture history, an initial posture reference, and object geometry.

\vspace{5pt}
\noindent \textbf{Problem Formulation.}
We focus on posture- and object-aware co-speech gesture generation. Given an $H$-frame gesture motion history $\textbf{h}\in\mathbb{R}^{H\times D}$, speech signal $\mathcal{S}$ containing audio and text, an initial posture reference $\gamma$ and surrounding 3D object mesh $\mathcal{O}$, the goal is to autoregressively generate continuous and realistic gesture sequences $\mathbf{G} \in \mathbb{R}^{F \times D}$. 
To support long-horizon generation, the gesture sequence is decomposed into gesture primitives $\mathbf{G} = \{\mathbf{g}^{(1)}, \dots, \mathbf{g}^{(K)}\}$. Our goal is to synthesize primitives that align rhythmically and semantically with speech, smoothly transition from preceding history, preserve the referenced posture, and satisfy the physical constraints of surrounding objects. 
To achieve this, we first model the core communicative motion using a canonical, shape-agnostic body representation to learn a strong speech-gesture prior (Stages 1 and 2). We subsequently inject specific body shape parameters and object geometries to spatially ground the gestures within the physical environment (Stage 3).


\subsection{Stage1: Causal Latent Gesture Primitive Space}

To efficiently model long-horizon co-speech gestures without the computational burden of diffusion in high-dimensional motion space or the quantization artifacts of discrete tokenization, we operate in a continuous latent space. 
Unlike bidirectional latent models that obscure temporal structure and limit control over speech-motion alignment, we adopt a \emph{causal} VAE for gesture modeling that enforces temporal ordering along the time axis, enabling more explicit control over speech–motion alignment. 
This provides a compact continuous space for diffusion while preserving audio–gesture alignment and supporting gesture completion and in-betweening. The causal structure also facilitates autoregressive diffusion in Stage~2.



\begin{figure*}[t]
    \centering
    \includegraphics[width=0.97\linewidth]{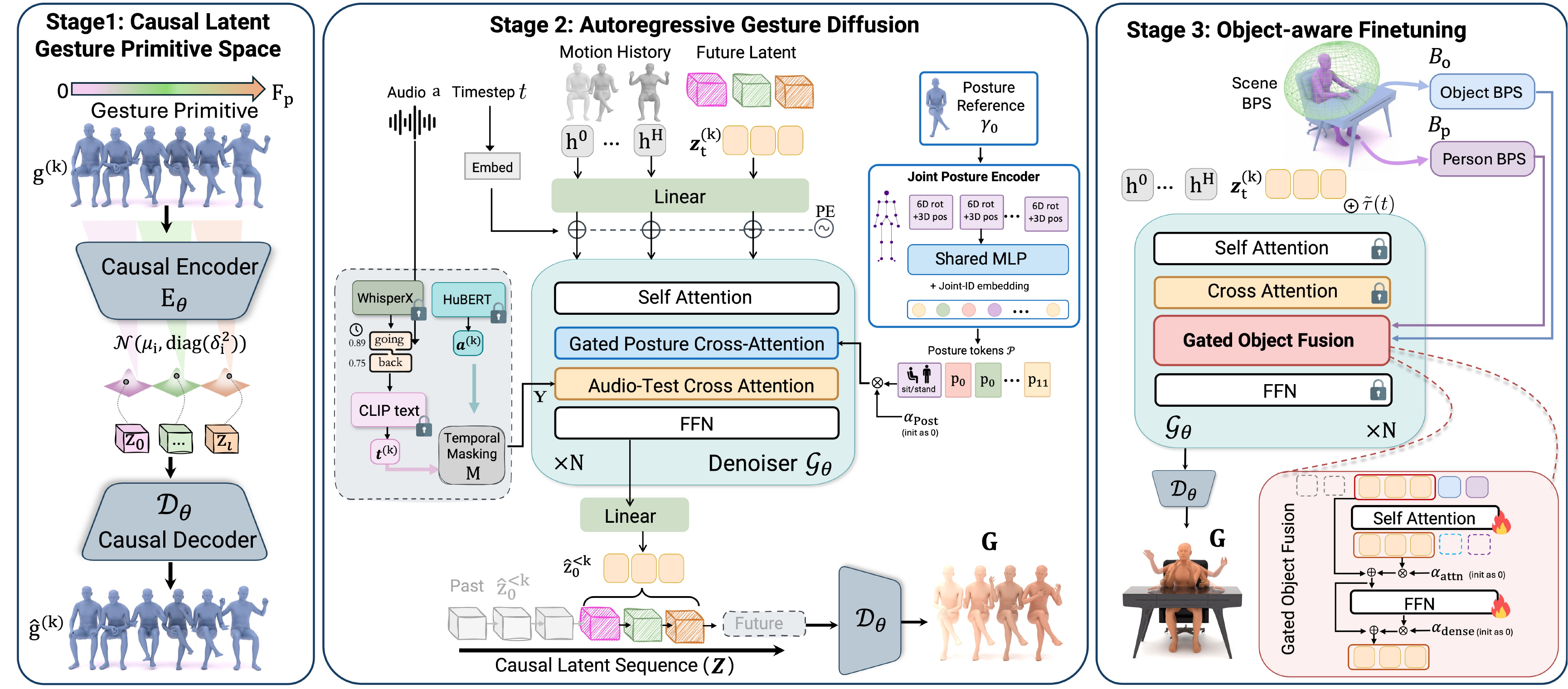} \vspace{-10pt}
    \caption{\footnotesize Overview of \puppet. (Stage1) A CausalVAE encodes gesture primitives into compact, temporally ordered latent tokens.
    (Stage2) Autoregressive diffusion operates in this latent space, using temporal masking to synchronize generated motions with speech and posture.
    (Stage3) A Gated Object Fusion module uses Basis Point Set (BPS) representations to inject scene awareness into the frozen base model, enabling physically grounded gestures.}
    \vspace{-10pt}
    \label{fig:main}
\end{figure*}

\vspace{5pt}
\noindent \textbf{Architecture.}
Given a gesture primitive $\mathbf{g}\in\mathbb{R}^{F_p\times D}$, where $F_p$ is the primitive length and $D$ the per-frame feature dimension, the encoder $E_{\phi}$ encodes $\mathbf{g}$ to the parameters of a temporally factorized Gaussian posterior with reduced temporal resolution to get temporally ordered latent tokens:
\begin{equation}
\{(\boldsymbol{\mu}_1, \boldsymbol{\sigma}_1^2), \dots, (\boldsymbol{\mu}_l, \boldsymbol{\sigma}_l^2)\}=E_{\phi}(\mathbf{g}),
\qquad
\boldsymbol{\mu}_i, \boldsymbol{\sigma}_i^2 \in \mathbb{R}^{d}
\end{equation}
where $l=F_p / s$ with encoder’s temporal downsampling factor $s$, and $d$ as the latent dimensionality.
The posterior factorizes causally across latent tokens:
\begin{equation}
\resizebox{0.90\columnwidth}{!}{$
q_{\phi}(\mathbf z\mid\mathbf g)
= \prod_{i=1}^{L} q_{\phi}(\mathbf z_i\mid\mathbf g_{\le i}),
\quad
q_{\phi}(\mathbf z_i\mid\mathbf g_{\le i})
= \mathcal N(\boldsymbol\mu_i,\operatorname{diag}(\boldsymbol\sigma_i^2))
$}
\end{equation}
where $\mathbf{g}_{\le i}$ denotes the encoder’s causal receptive field for token $i$.
The latent tokens $\mathbf{z}_i \in \mathbb{R}^{d}$, are obtained via reparameterization,
($
\mathbf{z}_i 
= \boldsymbol{\mu}_i 
+ \boldsymbol{\sigma}_i \odot \boldsymbol{\epsilon}_i,
\;
\boldsymbol{\epsilon}_i \sim \mathcal{N}(\mathbf{0}, \mathbf{I})
$),
yielding $\mathbf{z} = \{\mathbf{z}_1,\dots,\mathbf{z}_l\}\in\mathbb{R}^{l\times d}$.
The encoder uses 1D causal convolutions with strided temporal downsampling and dilated residual blocks. 
Causality restricts each latent token to past and current frames within its receptive field.
The decoder $D_{\theta}$ mirrors this causal architecture with temporal upsampling, while a learned projection $\psi$ aligns the latent dimensionality with the decoder feature width:
\begin{equation}
\tilde{\mathbf{z}} = \psi(\mathbf{z}),
\qquad
\hat{\mathbf{g}} = D_{\theta}(\tilde{\mathbf{z}}),
\qquad
\hat{\mathbf{g}} \in \mathbb{R}^{F_p \times D}.
\end{equation}
This yields a compact, temporally structured latent sequence that preserves causal ordering for autoregressive diffusion in Stage~2. 
The CausalVAE is trained with feature reconstruction, auxiliary temporal consistency, KL regularization, and SMPL-X–based geometric consistency losses. Full details are provided in Appendix~\ref{app:cvae}.



\subsection{Stage 2: Autoregressive Gesture Diffusion}
\label{sec:stage2}

Building on the compact casual latent primitive space, we model co-speech gesture generation as autoregressive diffusion over latent primitives conditioned on speech signals (audio and text), gesture history, and an initial posture reference.
Let $\mathbf{g}^{(k)} \in \mathbb{R}^{F_p \times D}$ be the $k$-th gesture primitive and $\mathbf{z}_0^{(k)}\in \mathbb{R}^{l \times d}$ its latent tokens from $E_{\phi}$.
A long sequence is represented as an ordered sequence of latent primitives $\mathcal{Z}_{1:K} = \{\mathbf{z}_0^{(1)}, \dots, \mathbf{z}_0^{(K)}\}$, with conditional distribution
\begin{equation}\vspace{-5pt}
p_{\theta}\big(\mathcal{Z}_{1:K} \mid \mathbf{h}, \mathbf{\mathcal{S}}, \gamma\big)
= \prod_{k=1}^{K}
p_{\theta}\big(\mathbf{z}^{(k)} \mid \mathbf{h}^{(k)}, \mathcal{S}^{(k)},\gamma\big),
\label{eq:ar_factorization_hist}
\end{equation}
where $\mathbf{h}^{(k)} \in \mathbb{R}^{H \times D}$ is the last $H$ motion frames preceding primitive $k$ and $\mathcal{S}^{(k)} = \{\mathbf{a}^{(k)}, \mathbf{t}^{(k)}\},$ contains audio features $\mathbf{a}^{(k)} \in \mathbb{R}^{F_a \times D_a}$ and token-level text features $\mathbf{t}^{(k)} \in \mathbb{R}^{F_t \times D_t}$ of primitive $k$. 
The initial posture reference $\gamma$ is fixed for the full sequence and reused across all autoregressive primitives.
We omit the superscript $(k)$ hereafter.

\noindent\textbf{Latent diffusion over gesture primitives.}
For each gesture primitive (i.e., each factor in~\cref{eq:ar_factorization_hist}), we define a forward diffusion process 
$q(\mathbf{z}_t \mid \mathbf{z}_0)
= \mathcal{N}\big(\sqrt{\bar{\alpha}_t}\mathbf{z}_0,\ (1-\bar{\alpha}_t)\mathbf{I}\big)$ 
over $T$ timesteps where $\bar{\alpha}_t = \prod_{s=1}^{t} \alpha_s$ and $\{\alpha_t\}_{t=1}^{T}$ is a fixed variance schedule.
We train a conditional diffusion model $\mathcal{G}_\theta$ to predict the clean latent $\hat{\mathbf{z}}_0=\mathcal{G}_\theta\!\left(\mathbf{z}_t,\; t,\; \mathbf{h},\; \mathcal{S}\right)$ with $\mathbf{z}_t = \sqrt{\bar{\alpha}_t}\mathbf{z}_0 + \sqrt{1-\bar{\alpha}_t}\boldsymbol{\epsilon}$.
The denoiser is trained with a motion-space $g_0$ loss and a latent consistency loss, $\mathcal{L}=\lambda_{g_0}\,\mathcal{L}_{g_0}+\lambda_{\text{latent}}\,\mathcal{L}_{\text{latent}}$.
A uniform MSE weights all frames equally, despite high-intensity gestures being more perceptually salient and difficult to model than near mean-pose motion. To emphasize dynamic regions, we introduce an intensity-weighted loss:
{\small
\begin{equation}
\mathcal{L}_{g_0}
=
\mathbb{E}_{\mathbf{g}_0}
\left[
\frac{1}{Z_\omega}
\sum_{f=1}^{F}
\omega_f
\left\|
\mathbf{g}_0^{(f)}-\hat{\mathbf{g}}_0^{(f)}
\right\|_2^2
\right].
\end{equation} 
}
where $\mathbf{g}_0$ and $\hat{\mathbf{g}}_0$ are the ground-truth and generated motion features decoded from the latent, and $\omega_f = 1 + \lambda_I I_f^{2}$ is the frame-wise weight with normalization factor $Z_\omega = \sum_{f=1}^{F}\omega_f$.
Here, $I_f$ is the normalized motion intensity at frame $f$, and $\lambda_I$ controls reweighting strength. 
This preserves supervision over low-intensity motion while allocating greater gradient emphasis to high-energy gesture segments.
We also enforce consistency between predicted and ground-truth primitive latents in the causal space:
\begin{equation}
\mathcal{L}_{\text{latent}}
=
\mathbb{E}_{\mathbf{z}_0}
\left[
\left\|
\mathbf{z}_0
-
\hat{\mathbf{z}}_0
\right\|_2^2
\right].
\end{equation}
During sampling, we apply classifier-free guidance~\cite{ho2022classifier}:
\begin{equation}
\begin{aligned}
\mathcal{G}_w(\mathbf{z}_t,&t,\mathbf{h},\mathcal{S},\gamma)
={}
\mathcal{G}_\theta(\mathbf{z}_t,t,\mathbf{h},\varnothing,\gamma)
\\
&+ w\Big[
\mathcal{G}_\theta(\mathbf{z}_t,t,\mathbf{h},\mathcal{S},\gamma)
-
\mathcal{G}_\theta(\mathbf{z}_t,t,\mathbf{h},\varnothing,\gamma)
\Big].
\end{aligned}
\end{equation}
At inference, we sample $\mathbf{z}_T \sim \mathcal{N}(\mathbf{0}, \mathbf{I})$ and use DDIM~\cite{song2020denoising} for $N$ steps to obtain $\hat{\mathbf{z}}_0$,  which is decoded by the pretrained causal decoder as $\hat{\mathbf{g}} = D_{\theta}\!\left(\hat{\mathbf{z}}_0\right)$. 


\vspace{5pt}
\noindent \textbf{Denoiser Architecture.}
The denoiser concatenates (i) motion-space history embeddings and (ii) noisy future latent embeddings:
\begin{equation}
\mathbf{X} = \mathrm{PE}\left( \left[ \mathbf{W}_h \mathbf{h},\ \mathbf{W}_z \mathbf{z}_t \right] \right) + \tau(t) \in \mathbb{R}^{(H+l)\times d_e}, \label{eq:token_concat}
\end{equation}
where $\mathbf{W}_h\in\mathbb{R}^{D\times d_e}$ and $\mathbf{W}_z\in\mathbb{R}^{d\times d_e}$ are linear projections, $\mathrm{PE}(\cdot)$ is positional encoding, and $\tau(t)$ is the diffusion timestep. Each transformer block applies self-attention over all $(H+l)$ tokens to propagate history into denoising.

Since a short history window may not preserve fine-grained posture over long autoregressive generations, we also condition on the initial posture reference $\gamma$, taken from the first frame and fixed throughout generation.
A joint-level posture encoder maps the lower-body to posture tokens $\mathbf{P}$, augmented with a learned sitting/standing token as a coarse posture cue. 
Cross-attention is applied only to future latent tokens ($\mathbf{X}_{\text{fut}} = \mathbf{X}_{H+1:H+l}$).
Each transformer block injects posture through gated cross-attention:
\begin{equation} 
\mathbf{X}_{\mathrm{fut}} \leftarrow \mathbf{X}_{\mathrm{fut}} + \tanh(\alpha_{\mathrm{post}}) \mathrm{CrossAttn} (\mathbf{X}_{\mathrm{fut}},\mathbf{P}), \label{eq:posture_attn} 
\end{equation}
where $\alpha_{\mathrm{post}}$ is a learnable scalar initialized to zero, gradually introducing posture conditioning while preserving the base denoising path at initialization.
Speech conditioning is also applied to future latent tokens.  
Let $\mathbf{Y}_a = \mathbf{W}_a\mathbf{a} \in\mathbb{R}^{F_a\times d_e}$ and $\mathbf{Y}_t=\mathbf{W}_t\mathbf{t} \in\mathbb{R}^{F_t\times d_e}$ be projected audio and text conditions with $\mathbf{Y}=[\mathbf{Y}_a;\mathbf{Y}_t]$. Each block updates as:
\begin{equation}
\mathbf{X}_{\text{fut}} \leftarrow \mathbf{X}_{\text{fut}} + \mathrm{CrossAttn}(\mathbf{X}_{\text{fut}}, \mathbf{Y}; \mathbf{M}),
\label{eq:crossattn}
\end{equation}
where $\mathbf{M}$ is a temporally controlled cross-attention mask.

\vspace{5pt}
\noindent\textbf{Temporal Control via Adaptive Masking.}
Leveraging the explicit temporal ordering in the causal latent space, we further introduce two alignment mechanisms to improve gesture–speech synchronization. First, an \emph{Audio Window Attention} mechanism restricts each latent index $i$ to attend to a local audio neighborhood centered at its temporally aligned audio position, promoting rhythmic consistency. We introduce a scaling coefficient $\xi$ that controls the size of this temporal attention window, defined as $w_a = \xi \frac{F_a}{l}$.Second, a \emph{Text Span Attention} mechanism constrains latent indices that fall within the temporal spans of text tokens to attend primarily to the corresponding text token, reinforcing semantic alignment. Both mechanisms are enforced via the cross-attention mask $\mathbf{M}$ in~\cref{eq:crossattn}, restricting each latent index to an adaptive conditioning window. 
Mask construction details are given in Appendix~\ref{app:latent_denoiser}.
Finally, a linear projection maps $\mathbf{X}_{\text{fut}}$ to the predicted clean latent tokens 
$\hat{\mathbf{z}}_0$.





\subsection{Stage3: Object-Aware Gesture Generation} \vspace{-5pt}
\label{sec:stage3}

To preserve the gesture prior 
while adding object awareness, we freeze all Stage~2 components and introduce a gated object fusion module as an additional conditional branch.

\vspace{5pt}
\noindent \textbf{Object and Shape Representation.}
At each denoising step $t$, we use Basis Point Sets (BPS)~\cite{prokudin2019efficient} to represent (i) the person at the last history frame and (ii) surrounding objects.
Unlike the original BPS that employs a unit sphere, we use an ellipsoidal support with larger lateral and frontal radii ($r_{\text{lat}}, r_{\text{front}}$) than vertical radius ($r_{\text{vert}}$), reflecting that most collisions arise from lateral and forward hand interactions with nearby furniture. 
We randomly sample and fix 1024 BPS points within this support.
During generation, $P$ points are uniformly sampled from the surrounding meshes, and the minimum distance from each BPS point to these samples forms $B_O \in \mathbb{R}^{1024}$.
For the person, we use linear blend skinning (LBS) weights~\cite{loper2015smpl} to select upper-body 
vertices and compute the minimum distance from these vertices to each BPS point, forming $B_P \in \mathbb{R}^{1024}$. 
The combined representation $\mathbf{B}_\text{H} = [B_O, B_P]$ represents the human-object relationship at the start of each gesture primitive.

\vspace{5pt}
\noindent \textbf{Gated Object Fusion.}
To inject object awareness while preserving the pretrained gesture prior, we insert a Gated Object Fusion module between the self-attention and speech cross-attention:
\begin{equation}
\begin{aligned}
\mathbf{X}_{\text{fut}} &\leftarrow \mathbf{X}_{\text{fut}} + \mathrm{SelfAttn}(\mathbf{X}_{\text{fut}}), \\
\mathbf{X}_{\text{fut}} &\leftarrow \mathrm{ObjFuser}(\mathbf{X}_{\text{fut}}, \mathbf{B}_{\text{H}}), \\
\mathbf{X}_{\text{fut}} &\leftarrow \mathbf{X}_{\text{fut}} + \mathrm{CrossAttn}(\mathbf{X}_{\text{fut}}, \mathbf{Y}; \mathbf{M}).
\end{aligned}
\label{eq:object-fusion}
\end{equation}
The fusion module follows a gated transformer design~\cite{li2023gligen}:
\begin{align}
\mathbf{X}_{\text{fut}}
&= \mathbf{X}_{\text{fut}}
+ \tanh\!\big(\alpha_{\text{attn}}\big)\;
\mathrm{TS}\!\left(
\mathrm{SelfAttn}\!\left(
[\mathbf{X}_{\text{fut}},\mathbf{B}_{\text{H}}]
\right)\right)
\notag \\
\mathbf{X}_{\text{fut}}
&= \mathbf{X}_{\text{fut}}
+ \tanh\!\big(\alpha_{\text{dense}}\big)\;
\mathrm{FeedForward}\!\left(\mathbf{X}_{\text{fut}}\right).
\label{eq:fut_update}
\end{align}
where $\mathrm{TS}(\cdot)$ retains only future gesture tokens, 
$\alpha_{\text{attn}}$ and $\alpha_{\text{dense}}$ are learnable scalars initialized to $0$. To further preserve the Stage~2 gesture prior during object-aware training, we interleave object-training batches with BEAT2 and Embody3D replay samples at a $4{:}1$ ratio. Replay samples use a learned no-object token to represent absent object geometry.

\noindent \textbf{Training.} We optimize a weighted sum of three losses:
\[
    \mathcal{L} = \lambda_{x_0} \mathcal{L}_{x_0} + \lambda_{latent} \mathcal{L}_{latent} + \lambda_{collision} \mathcal{L}_{collision},
\]
where $\lambda_{x_0} = \lambda_{latent} = 1$, and $\lambda_{collision} = 50$. Replay samples use only $\mathcal{L}_{x_0}$ and $\mathcal{L}_{latent}$, without collision supervision. Details are provided in Appendix~\ref{app:stage3}.

\section{\sceneges{} Dataset}

\begin{figure*}[t]
    \centering
    \includegraphics[width=0.9\linewidth]{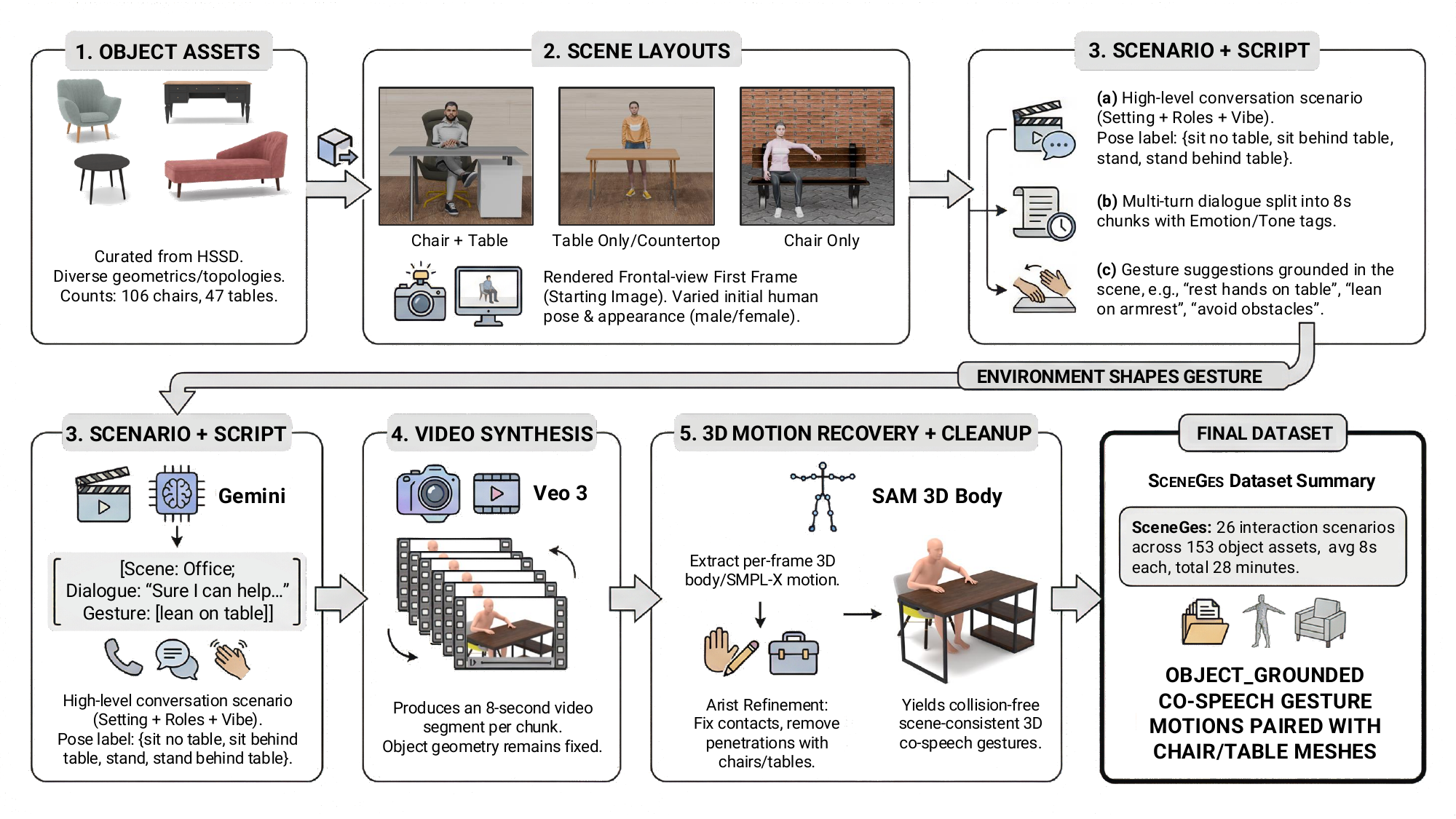}  \vspace{-20pt}
    \caption{\footnotesize Pipeline for constructing \sceneges{} dataset. Objects are arranged into scene layouts, conversation scenarios and gesture scripts are generated with Gemini, videos are synthesized with Veo~3, and SMPL-X motion is recovered and refined to produce collision-free, object-grounded co-speech gestures.}
    \vspace{-15pt}
    \label{fig:sceneges_pipeline}
\end{figure*}

Existing gesture datasets primarily capture speech-driven motions in standing or controlled studio environments without modeling surrounding objects. In practice, gestures are shaped by the environment; people rest their hands on tables, lean on armrests, and adapt movements to nearby furniture. Without object representations, models cannot learn such object-grounded behaviors.

To address this, we introduce \sceneges, a synthetic dataset for scene-aware co-speech gesture generation that pairs diverse chair and table assets with 3D SMPL-X~\cite{pavlakos2019expressive} gesture motions. As shown in Fig.~\ref{fig:sceneges_pipeline}, we curate object assets, construct simple scene layouts, generate scenario-driven dialogue and gesture scripts, synthesize scene-conditioned videos, and recover 3D motion that is further refined to ensure physically plausible object interactions.


\sceneges{} contains 26 interaction scenarios instantiated across $153$ object assets ($106$ chairs and $47$ tables). Each sequence has an average duration of $8$ seconds, yielding $28$ minutes of object-grounded co-speech motion covering diverse conversational contexts and object interactions. 
We further validate \sceneges{} through a user study and kinematic comparison with Embody3D, showing that \sceneges{} is perceptually difficult to distinguish from real motion and closely matches real gesture statistics. Additional dataset and validation details are provided in Appendix~\ref{app:db}.

\section{Experiments}
\label{sec:experiments}

\noindent\textbf{Datasets.}
We train and evaluate our model on BEAT2~\cite{liu24emage} and Embody3D~\cite{mclean2025embody} and add object awareness with our proposed \sceneges{} dataset. BEAT2 contains SMPL-X gestures paired with audio from 25 speakers, all recorded standing while reading predefined transcripts in relatively limited scenarios. Embody3D includes 280 speakers in dyadic and multi-person conversations, captured in sitting and standing modes across a broader range of everyday conversational scenarios.

\vspace{5pt}
\noindent\textbf{Metrics.} Following prior work~\cite{liu24emage,chen2025language,liu2025gesturelsm}, we report FGD~\cite{yoon2020speech} for gesture realism, and BC~\cite{li2021ai} for speech-motion synchrony. 
Recent studies~\cite{yang2025gesturehydra, chen2024diffsheg, cheng2024siggesture} note that these metrics do not fully reflect the perceptual quality.
Therefore, we also report $\Delta$BC~\cite{yang2025gesturehydra}, the average absolute difference between ground-truth and generated BC, measuring how closely the generated gestures match the synchrony of the ground-truth speaker. 
We further introduce inter-sample gesture diversity (Div), measuring variation among multiple gestures generated for the same audio. Specifically, we compute the average pairwise L$_1$ distance between upper-body joint positions across generated samples, normalized by the temporal median of each joint coordinate. This normalization prevents gestures with meaningful but low-intensity motion from being unfairly penalized, and better captures semantic diversity rather than favoring large-amplitude motion.
To evaluate posture accuracy, we compute the per-frame L$_1$ distance between ground-truth and generated lower-body joint positions (LL1). 
For human-object collisions, we compute body-vertex penetration depth inside surrounding objects using a signed distance field, normalized by the BPS ellipsoid extent. For each sample, we report maximum (MaxPen) and mean (MeanPen) penetration across all frames and vertices, then take the median across samples.

\subsection{Quantitative Evaluation}
\begin{table}[t]
\setlength{\tabcolsep}{8pt}
    \centering \scriptsize
    \caption{\footnotesize Baseline comparison for co-speech gesture generation on the BEAT2 (Speaker2) test set. All metrics are reported in  $\times 10^{-1}$.}
    \vspace{-0.1in}
    \begin{tabular}{lcccc}
        \toprule
        Methods 
        & FGD$\downarrow$ & BC$\uparrow$ & $\Delta$BC$\downarrow$ & Div$\uparrow$\\
        \midrule
        DisCo~\cite{liu2022disco}           & 7.586 & 7.652         & 0.224 & N/A \\
        CaMN~\cite{liu22beat}               & 6.273 & 7.458         & \textbf{0.145} & N/A \\
        EMAGE~\cite{liu24emage}             & 5.117 & 5.910         & 1.522 & N/A \\
        EMAGE*~\cite{liu24emage}            & 5.205 & 6.048         & 1.377 & \underline{12.830} \\
        SynTalker~\cite{chen2024Synerg}     & 4.069 & 7.407         & 0.481 & 10.676 \\
        LOM~\cite{chen2025language}         & 4.538 & 6.114 & 1.318 & 9.112 \\
        MIBURI~\cite{mughal2026miburi}      & 4.369 & \underline{7.685}          & 0.369 & 9.645 \\
        GestureLSM~\cite{liu2025gesturelsm} & \underline{3.692} & 7.547          & \underline{0.166} & 9.490 \\
        \midrule
        \puppet                    & \textbf{3.436} & \textbf{7.693} &  0.201 & \textbf{14.131} \\
        \bottomrule
    \end{tabular}
    \vspace{-0.1in}
    \label{tab:sota_compare}
\end{table}

\begin{table}[t]
\centering
\scriptsize
\setlength{\tabcolsep}{1pt}
\renewcommand{\arraystretch}{0.9}
\caption{Training composition and posture conditioning effects.}\vspace{-10pt}
\begin{adjustbox}{max width=\linewidth}
\begin{tabular}{l c cc ccc}
\toprule

\multirow{3}{*}[-2.2ex]{\parbox[c]{18mm}{\centering Puppeteer\\Train Data}}
&
\multirow{3}{*}[-2.2ex]{\parbox[c]{14mm}{\centering Posture}}
& \multicolumn{5}{c}{Test Data} \\
\cmidrule(lr){3-7}

& &
\multicolumn{2}{c}{\parbox[c]{24mm}{\centering BEAT2\\(ALL speakers)}}
& \multicolumn{3}{c}{Embody3D}\\
\cmidrule(lr){3-4}
\cmidrule(lr){5-7}

& & FGD & Div &  FGD & Div & LL1  \\
\midrule

BEAT2~(Speaker2)        & N/A     & 15.402 & 13.993 & 33.649 & 14.074 & 16.606 \\
BEAT2~(All)          & N/A     & 3.550 & 12.062  & 7.543 & 11.272 & 16.409 \\
\midrule
BEAT2~(All)+Embody3D  & \xmark  & 3.433 & 11.978 &  2.458 & 11.909 & 5.553  \\
BEAT2~(All)+Embody3D  & modulation  & \underline{2.786} & \underline{12.086} & \underline{2.300} & \underline{12.021} &  \underline{4.297} \\
BEAT2~(All)+Embody3D  & cross-attn  & \textbf{2.769} & \textbf{12.116} & \textbf{1.676} & \textbf{12.174} &  \textbf{2.446} \\

\bottomrule
\end{tabular} \vspace{-20pt}
\end{adjustbox}
\label{tab:test_sets}
\end{table}









\noindent\textbf{Quantitative Comparison.}
\cref{tab:sota_compare} compares methods on the BEAT2 (Speaker2) test set. 
Our method achieves the best FGD, BC, and Div, indicating more realistic, synchronized, and diverse gestures, while remaining competitive in $\Delta$BC.
Prior work often measures diversity using an L1 diversity metric as the L1 distance between a generated motion and its mean pose (intra-sample), which tends to favor exaggerated motions over semantically distinct gestures.Instead, we compute diversity from pairwise distances between gestures generated from the same audio input, better capturing multiple plausible gesture realizations.
EMAGE~\cite{liu24emage} uses VQ-VAE masked generation and selects the most likely codebook index via argmax, producing deterministic outputs with no diversity. We instead sample codebooks from the softmax distribution to enable diverse generation (EMAGE\textbf{*}). While EMAGE* introduces diversity, it increases FGD, whereas other methods reduce FGD at the cost of lower diversity. Our method achieves both higher diversity and lower FGD, demonstrating a better balance between realism and diversity.

\noindent\textbf{Effect of Training Data and Posture Conditioning.} 
\cref{tab:test_sets} evaluates training data composition and posture conditioning across different test sets. Training only on BEAT2 performs well in-domain but generalizes poorly to Embody3D due to the domain gap. Adding Embody3D improves performance on both, indicating stronger cross-dataset generalization. As a simple posture baseline, we add a learned binary sitting/standing embedding to the diffusion timestep (modulation; see Appendix~\ref{app:latent_denoiser}), which further improves performance, particularly reducing LL1. 
Replacing this coarse conditioning with our posture reference cross-attention further reduces Embody3D FGD to 1.676 and LL1 to 2.446. This shows the benefit of modeling fine-grained posture, allowing the model to better capture posture-specific gesture patterns. 
Qualitative examples illustrating the effect of the posture conditioning are provided in the appendix.



\begin{table}[t]
\centering
\scriptsize
\caption{Object-awareness evaluation. Since no baseline directly supports object-grounded co-speech gesture generation, we use LoM as a proxy by placing its generations in the same structured scenes. MeanPen is reported in $\times10^{-4}$ and MaxPen in $\times10^{-1}$.}
\vspace{-10pt}
\setlength{\tabcolsep}{2.2pt}

\begin{tabular}{l|cccc}
\toprule
 & \parbox[c]{15mm}{\centering MeanPen $\downarrow$\\($\times10^{-4}$)}
 & \parbox[c]{15mm}{\centering MaxPen $\downarrow$\\($\times10^{-1}$)}
 & LL1$\downarrow$
 & Div$\uparrow$\\
\midrule
LoM in scene (proxy) & 12.282 & 1.465 & 6.563 & -- \\
\midrule
w/o obj module & 6.476 & 1.239 & 1.019 & 10.009 \\
w/ obj module, w/o $\mathcal{L}_{collision}$ & 7.302 & 1.090 & \textbf{0.760} & 12.061 \\
w/ obj module, w/ $\mathcal{L}_{collision}$ & \textbf{4.450} & \textbf{0.713} & 0.780 & 11.393 \\
\bottomrule
\end{tabular}
\vspace{-15pt}
\label{tab:abl-collision}
\end{table}
\noindent \textbf{Object Awareness.} 
\revise{Since no prior method supports object-grounded co-speech gesture generation with aligned 3D scene geometry, we use LoM as a proxy by placing its generations in the same structured scenes. As shown in Tab.~\ref{tab:abl-collision}, LoM produces substantially higher spatial and posture errors ($2.8\times$ higher MeanPen, $2.0\times$ higher MaxPen, and $8.4\times$ higher LL1) than our object-aware model, highlighting the gap between scene-agnostic gesture realism and object-grounded spatial plausibility.

Beyond this proxy comparison, we ablate the object-aware components. }
\cref{tab:abl-collision} shows, adding the object module improves lower-body accuracy and reduces unrealistic body–object interactions, significantly lowering LL1. 
Without object awareness, the character maintains its sit/stand posture but lacks cues from surrounding objects, often causing leg intersections with tables or chairs and increasing LL1 as shown in~\cref{fig:qualitative-3d}. 
When the object module is added without $\mathcal{L}_{collision}$, the model learns to leverage surrounding objects but lacks an explicit constraint to prevent intersections, leading to higher average body–object penetration. Incorporating $\mathcal{L}_{collision}$ substantially reduces MeanPen and MaxPen while maintaining similar LL1 and Div.
\revise{We further evaluate on a held-out object split, showing generalization to unseen furniture (see Appendix~\ref{app:unseen_obj}). }

\begin{table}[t]
\centering
\scriptsize
\setlength{\tabcolsep}{8pt}
\caption{\small Audio and text span masking ablation.}\vspace{-10pt}
\begin{tabular}{l c | c c c}
\toprule
Audio & Text & FGD$\downarrow$ & $\Delta$BC$\downarrow$ & Div$\uparrow$ \\
\midrule

Full   & --             & 3.917 & 0.431 & \underline{10.994} \\
Full   & Span           & 3.633 & \underline{0.273 }&  10.844 \\
\cline{1-5}
Window ($\xi=1$) \rule{0pt}{2.8ex} & --   & \textbf{3.385} & 0.346 & 10.462 \\
Window ($\xi=2$) & --   & 3.767 & 0.351 & 10.864 \\
Window ($\xi=3$) & --   & 3.779 & 0.348 & 10.815 \\
\cline{1-5}
Window ($\xi=1$) \rule{0pt}{2.8ex} & Span & \underline{3.437} & \textbf{0.201} & \textbf{14.131} \\

\bottomrule
\end{tabular}\vspace{-5pt}
\label{tab:audio_text_masking}
\end{table}

\begin{figure*}[t]
    \centering
    \includegraphics[width=0.89\linewidth]{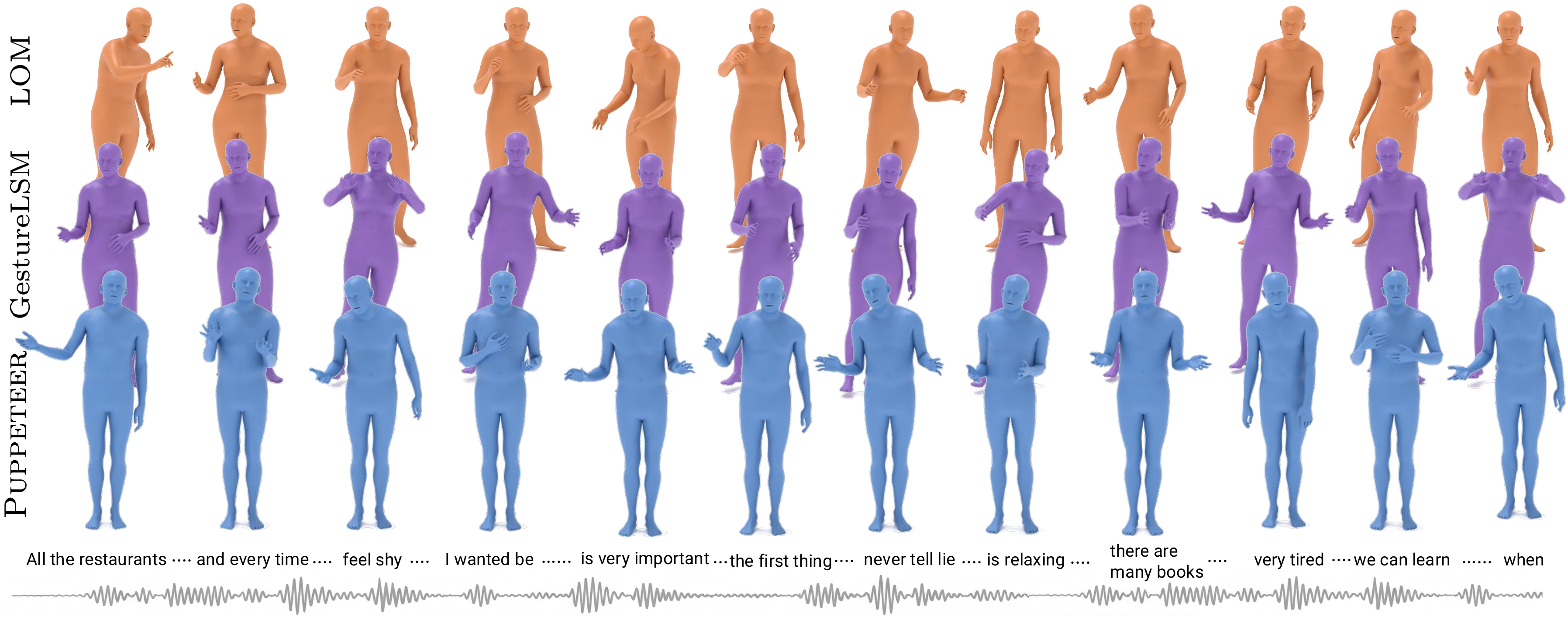} \vspace{-10pt}
    \caption{\small Qualitative comparison. \puppet{} generates more natural, speech-aligned gestures than LOM~\cite{chen2025language} and GestureLSM~\cite{liu2025gesturelsm}.}
    \vspace{-15pt}
    \label{fig:qualitative-comparison}
\end{figure*}

\subsection{Ablation Studies} \vspace{-3pt}





\begin{table}[t]
\centering \scriptsize
\setlength{\tabcolsep}{3pt}
\caption{\small Reconstruction results for VAE and CausalVAE.}\vspace{-10pt}
\begin{tabular}{c l l c c c}
\toprule
Train & Test & Methods & FGD$\downarrow$ & MPJPE$\downarrow$ & ACCL$\downarrow$ \\
\midrule

\multirow{4}{*}{
\begin{tabular}{c}
BEAT2 \\
Only
\end{tabular}
}
& \multirow{2}{*}{BEAT2}
& VAE                       & 0.242 & 16.757 & 15.013 \\
& & CasualVAE               & \textbf{0.161} & \textbf{7.387} & \textbf{5.497} \\
\cline{2-6}
& \multirow{2}{*}{Embody3D}
& VAE  \rule{0pt}{2.8ex}    &  \colorbox{orange!40}{1.344} & \colorbox{orange!40}{112.542} & \colorbox{orange!40}{73.057} \\
& & CasualVAE               & \colorbox{orange!40}{1.173} & \colorbox{orange!40}{107.324} & \colorbox{orange!40}{69.344} \\
\midrule

\multirow{4}{*}{
\begin{tabular}{c}
BEAT2+ \\
Embody3D
\end{tabular}
}
& \multirow{2}{*}{BEAT2}
& VAE       & 0.221 & 17.609 & 15.832 \\
& & CasualVAE & \textbf{0.168} & \textbf{7.526} & \textbf{6.646} \\
\cline{2-6}
& \multirow{2}{*}{Embody3D}
& VAE  \rule{0pt}{2.8ex}     & 0.590 & 18.416 & 8.971 \\
& & CasualVAE & \textbf{0.153} & \textbf{9.381} & \textbf{5.522} \\

\bottomrule
\end{tabular}
\label{tab:causalvae-recon}
\end{table}






\begin{table}[t]
\centering \scriptsize
\setlength{\tabcolsep}{10pt}
\caption{\small Gesture generation results using VAE and CausalVAE latents. \revise{Both use the same autoregressive primitive generation.}}\vspace{-10pt}
\begin{tabular}{l cccc}
\toprule
Methods & FGD$\downarrow$ & BC$\uparrow$ & $\Delta$BC$\downarrow$ & Div$\uparrow$ \\
\midrule

Real      & - & 7.282 & -  & - \\
\cline{1-5}
VAE   \rule{0pt}{2.8ex}    & 4.221 & 7.570 &  0.596 & 10.849\\
CasualVAE & \textbf{3.437} &\textbf{7.693} &  \textbf{0.201} & \textbf{14.131}\\

\bottomrule
\end{tabular}\vspace{-10pt}
\label{tab:causalvae-gen}
\end{table}
\noindent \textbf{Effect of CausalVAE.} The \textit{reconstruction} results in~\cref{tab:latent-dim-causalvae-recon} show that CausalVAE consistently outperforms the standard VAE across training settings. When trained and tested on BEAT2, it significantly reduces FGD, MPJPE, and acceleration error (ACCL), indicating more accurate reconstruction. The advantage becomes more evident under cross-dataset evaluation. Under cross-dataset evaluation, both models degrade when trained only on BEAT2 and tested on Embody3D due to the domain gap ({\setlength{\fboxsep}{1pt}\colorbox{orange!40}{orange}}), but CausalVAE remains better across all metrics. When trained on both BEAT2 and Embody3D, CausalVAE again performs best, particularly on Embody3D where it substantially improves all three metrics. These results suggest that the causal latent representations capture better motion structure and generalize better across datasets, as BEAT2 alone provides limited posture variation and interactions for training gesture VAEs.
The \textit{generation} results~(\cref{tab:causalvae-gen}) further demonstrate the benefits of CausalVAE. Compared to VAE, it achieves lower FGD, higher BC, lower $\Delta$BC, and higher Div. Higher BC with lower $\Delta$BC indicates stronger speech–motion synchrony, with generated gestures more closely following speech rhythm and matching the synchrony of real gestures.
This improvement is enabled by the causal latent representation, which allows masking over text and audio conditioning during training and encourages stronger speech–motion relationships, improving audio–gesture alignment.
Higher Div indicates that CausalVAE produces a wider range of plausible gestures for the same speech input.

\noindent \textbf{Audio and Text Masking.} \cref{tab:audio_text_masking} evaluates different audio and text masking strategies. Full audio without masking serves as the baseline. 
Text span masking slightly improves FGD and substantially reduces $\Delta$BC, with a small drop in diversity.
Temporal audio cross-attention further improves results, with the best FGD at $\xi=1$; larger $\xi$ values widen the window and shift performance toward the Full setting, indicating weaker benefits from less localized conditioning. 
Combining the audio window with text span masking gives the best overall result, increasing diversity while further reducing the synchronization gap to ground truth.

\subsection{Qualitative Evaluation}

\noindent \textbf{Baseline Comparison.} As shown in~\cref{fig:qualitative-comparison}, \puppet{} better follows speech timing and emphasis than prior methods. For example, during phrases such as “I wanted to be …” and “the first thing …”, \puppet{} produces aligned self-referential hand-to-chest cues and brief index-finger emphasis gestures. 
LOM shows noticeable variation over time and some degree of speech synchronization, but gestures are less precisely aligned with specific speech segments. GestureLSM shows smoother orientation changes than LOM  but weaker correspondence between gesture timing and speech emphasis. 
Overall, \puppet{} better captures the temporal structure of speech and produces gestures that better match spoken phrases in timing and expressiveness.
whose timing and expressiveness more consistently match the spoken phrases.
The causal latent representation also enables motion in-betweening and completion; additional qualitative examples are provided in Appendix~\ref{app:qualitative}.

\begin{figure}[t]
    \centering
    \includegraphics[width=1\linewidth]{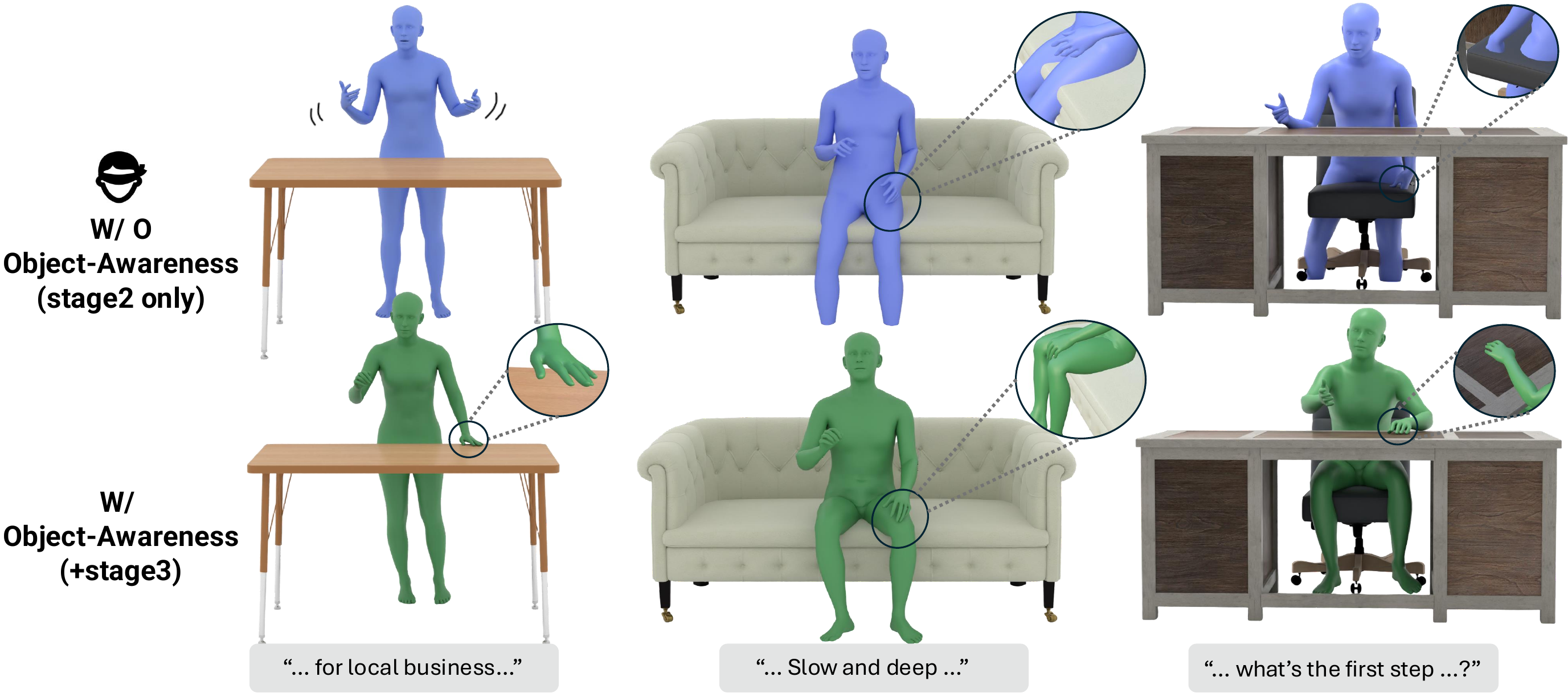} \vspace{-15pt}
    \caption{\small Qualitative comparison with and without object awareness. Stage2 (no object fusion) ignores nearby furniture, while adding Stage3 encourages object-grounded gestures.}\vspace{-15pt}
    \label{fig:qualitative-3d}
\end{figure} 
\noindent \textbf{Object Awareness.} \cref{fig:qualitative-3d} shows that Stage 3 enables gestures to adapt to surrounding furniture. Without object fusion (Stage2), motions often ignore nearby geometry, causing hand-table or leg-chair intersections. With object awareness (Stage3), the model produces physically grounded gestures, e.g., placing hands on the table while speaking and maintaining a sitting posture aligned with the chair.

\vspace{-10pt}
\section{Conclusion}
\vspace{-5pt}
We introduced \puppet{}, the first framework for object-grounded co-speech gesture generation, explicitly modeling the coupling between communicative gestures, posture, and surrounding objects. Our model operates in a causal latent space, enabling autoregressive generation with explicit temporal control over speech–gesture alignment. 
To support posture-aware and object-grounded gesture modeling, we introduce \sceneges{}, the first dataset for object-grounded co-speech gesture generation, together with new evaluation metrics. Experiments show improved realism, synchronization, and diversity over prior methods.


{
    \small
    \bibliographystyle{ieeenat_fullname}
    \bibliography{main}
}
\clearpage
\appendix
\setcounter{table}{0}
\setcounter{figure}{0}
\setcounter{page}{1}
\maketitlesupplementary


\renewcommand{\thesection}{\Alph{section}}

\renewcommand{\thetable}{S\arabic{table}}
\renewcommand{\thefigure}{S\arabic{figure}}

\noindent\textbf{Video.} In our supplemental video, an AI voiceover briefly introduces the motivation and the proposed approach. We then present qualitative comparisons between our method and prior approaches (LoM~\cite{chen2025language} and GestureLSM~\cite{liu2025gesturelsm}). 
Next, we illustrate the effect of posture conditioning on the generated gestures, examples of gesture in-betweening and completion, and object-aware gesture generation. We also include samples from our \sceneges{} dataset.
 We encourage the reader to watch the video to better appreciate the dynamic motion produced by our method.

\section{Gesture Primitive Representation} 
\label{app:gesture_repr}

\subsection{Representation}
We represent each frame as $\mathbf{y}_t = (\mathbf{R}_t,\; \mathbf{\Theta}_t,\; \mathbf{J}_t,\; \Delta \mathbf{R}_t,\; \Delta \mathbf{J}_t),$
where $\mathbf{R}_t \in \mathbb{R}^{6}$ denotes the 6D rotation~\cite{zhou2019continuity} of the global body orientation, $\mathbf{\Theta}_t \in \mathbb{R}^{(J-1)\times 6}$ denotes of $J-1=54$ joints,
$\mathbf{J}_t \in \mathbb{R}^{J\times 3}$ is the corresponding 3D joint locations obtained by SMPL-X forward kinematics~\cite{pavlakos2019expressive} under a \emph{canonical shape}, 
$\Delta \mathbf{R} \in \mathbb{R}^{6}$ denotes the 6D relative rotation of the global orientation between frames $(t-1,t)$, 
and $\Delta \mathbf{J}_t \in \mathbb{R}^{J\times 3}$ denotes the joint displacement $\mathbf{J}_t-\mathbf{J}_{t-1}$, yielding $D=666$.
A gesture primitive of length $F_p$ is then represented as \mbox{$\mathbf{g} = [\mathbf{y}_1,\dots,\mathbf{y}_{F_p}]\in\mathbb{R}^{F_p\times D}$}.

We represent rotations using the continuous 6D parameterization, to avoid the discontinuities of axis-angle parameterizations. Joint positions $\mathbf{J}_t$ are included in addition to rotations to provide an explicit Cartesian description of the articulated configuration derived through SMPL-X forward kinematics. This makes the representation over-parameterized but well-conditioned: rotations capture the pose in a form compatible with the body model, while positions expose the resulting spatial configuration that is directly relevant for gesture amplitude and hand trajectory modeling. Finally, the first-order differences $(\Delta \mathbf{R}_t, \Delta \mathbf{J}_t)$ make local temporal evolution explicit, allowing the encoder to model short-term dynamics without inferring them implicitly from consecutive frames. 

\subsection{Canonicalization}
\label{app:canonicalization}
\subsubsection{Neutralize the lower body} Large-scale gesture datasets sometimes include small body movements or trajectory shifts, even when the person is standing still. In downstream applications such as 3D animation, these unintended movements can cause problems, such as characters entering restricted areas or creating unwanted collisions. To prevent this, we remove trajectory during gesture generation in standing mode. We use inverse kinematics to keep the lower body fixed in a neutral position while allowing the upper body to move naturally. This constraint is applied only when the character is standing. Although we lock the lower body in standing mode, the model still generates full-body gestures and remains aware of overall body posture during generation. For example, different sitting postures can influence the generated gestures, meaning the model adapts its output depending on whether the character is sitting upright, leaning back, or in another seated configuration.
\subsubsection{Removing the shape from motion representation} The representation introduced in~\cref{app:gesture_repr} includes joint positions, which depend on both pose rotations and body shape. To ensure that the model learns only gesture-related information, we fix the body shape to the average shape (i.e., $\beta = 0$) for all gesture sequences during training. Although body shape is not encoded in the gesture primitive representation, it remains important in the final stage of training for collision avoidance. Specifically, when computing the BPS representation of the human body from SMPL-X vertices, we use the actual body shape parameters. By adjusting the SMPL-X vertices according to the corresponding $\beta$ values, we can more accurately account for body geometry and prevent collisions.

\section{Scene Representation}

\begin{figure*}[ht]
    \centering
    \includegraphics[width=0.85\linewidth]{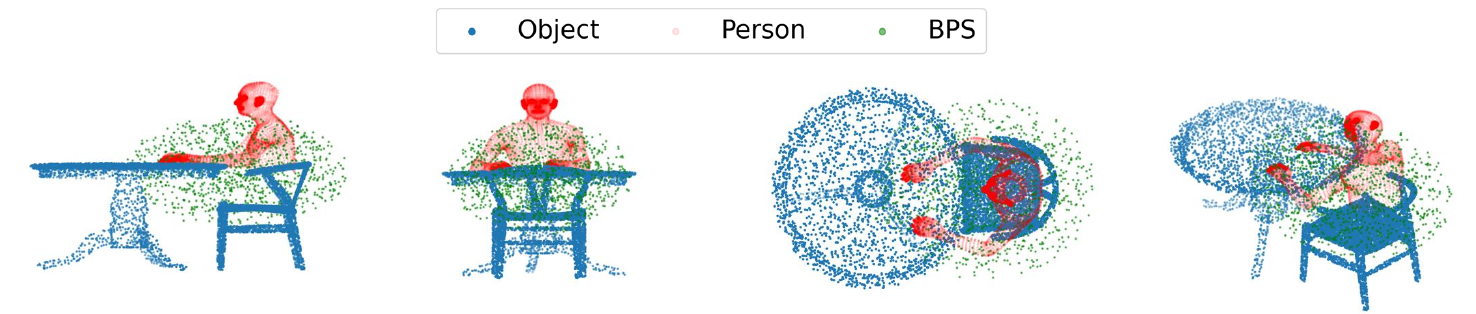}\vspace{-10pt}
    \caption{
    Scene representation illustrating the spatial relationship between the person, surrounding objects, and the BPS points. The person is shown in red, objects (table and chair) in blue, and BPS samples in green.
    }\vspace{-10pt}
    \label{fig:scene_representation}
\end{figure*}

As shown in~\cref{fig:scene_representation}, we use BPS points (shown in green) to obtain a compact representation of the spatial relationship between the person and the surrounding objects without explicitly modeling object topology. We define an ellipsoid with radii $(r_{\text{front}}, r_{\text{lat}}, r_{\text{vert}}) = (0.48, 0.6, 0.4)$ to cover a larger region in the frontal and lateral directions than in the vertical direction, since collisions are more likely to occur due to hand movements. The ellipsoid center is shifted by 0.5\,m forward and 0.1\,m upward to better capture interactions that predominantly happen in front of the person and around the upper body. 

Meshes often contain large flat regions that can be represented with relatively few vertices connected by triangular faces. As a result, computing minimum distances directly using mesh vertices can be unreliable. Following prior work~\cite{yi2024generating, yi2022human}, we uniformly sample $P = 3000$ points from the mesh surface. We then compute the minimum distance from each of the 1024 BPS points to these sampled points and store the resulting distances as $B_o \in \mathbb{R}^{1024}$.

For both scene representation and collision computation, we exclude the lower body. When sitting, the hip region is typically in contact with the chair surface, which can bias the collision loss. In practice, we observed that the model may exploit this by crossing the legs and penetrating the chair to reduce collision penalties, since collisions around the thighs produce a smaller loss than those at the hips. Instead, by relying on other loss terms, the model learns to maintain plausible lower-body configurations, as the lower body exhibits minimal motion in the training data during sitting.

\section{Gesture Causal VAE}
\label{app:cvae}

\noindent\textbf{Training.}
The CausalVAE is trained to reconstruct primitives while regularizing the latent posterior and enforcing temporal and geometric consistency using $\mathcal{L}_{\text{stage1}}$ loss:
\begin{equation}
\begin{aligned}
\mathcal{L}_{\text{stage1}}
= \lambda_{\text{rec}}\mathcal{L}_{\text{rec}}
+ &\lambda_{\text{kl}}\mathcal{L}_{\text{kl}}
+ \lambda_{\Delta j}\mathcal{L}_{\Delta j} 
+ \lambda_{\Delta r}\mathcal{L}_{\Delta r} \\
+ &\lambda_{\text{smpl}}\mathcal{L}_{\text{smpl}}
+ \lambda_{\text{cons}}\mathcal{L}_{\text{cons}}
\end{aligned}
\end{equation}
The reconstruction term is defined in the motion feature space as
\begin{equation}
\mathcal{L}_{\text{rec}} = \|\hat{\mathbf{g}} - \mathbf{g}\|,
\end{equation}
and the KL regularizer is computed over latent tokens: 
\begin{equation}
\mathcal{L}_{\text{kl}}
= \mathrm{KL}\!\left(q_{\phi}(\mathbf{z}\mid\mathbf{g}) \,\|\, \mathcal{N}(\mathbf{0},\mathbf{I})\right),
\end{equation}

To enforce internal kinematic consistency within the representation, we penalize discrepancies between the predicted velocity features and velocities \emph{recomputed} from reconstructed joints and rotations.
Specifically, let $\hat{\mathbf{J}}_t$ and $\hat{\mathbf{R}}_t$ denote the reconstructed joint positions and root rotations at time $t$. We recompute joint displacements and relative root rotations as
\begin{equation}
\widetilde{\Delta \mathbf{J}}_t 
= \hat{\mathbf{J}}_t - \hat{\mathbf{J}}_{t-1},
\qquad
\widetilde{\Delta \mathbf{R}}_t 
= \mathrm{Rel}(\hat{\mathbf{R}}_{t-1}, \hat{\mathbf{R}}_t),
\end{equation}
where $\mathrm{Rel}(\cdot,\cdot)$ denotes relative rotation expressed in 6D form.
Let $\widehat{\Delta \mathbf{J}}_t$ and $\widehat{\Delta \mathbf{R}}_t$ denote the velocity components directly predicted as part of the reconstructed feature vector. We define the consistency losses
\begin{equation}
\mathcal{L}_{\Delta j} 
= \|\widetilde{\Delta \mathbf{J}} - \widehat{\Delta \mathbf{J}}\|,
\qquad
\mathcal{L}_{\Delta r} 
= \|\widetilde{\Delta \mathbf{R}} - \widehat{\Delta \mathbf{R}}\|.
\end{equation}

To enforce geometric consistency, we recompute 3D joints via SMPL-X forward kinematics under canonical shape. 
Let $\mathcal{F}(\cdot)$ denote the SMPL-X forward operator mapping rotation parameters to joint positions. 
Given ground-truth global and local rotations $(\mathbf{R}, \mathbf{\Theta})$ and their reconstructions $(\hat{\mathbf{R}}, \hat{\mathbf{\Theta}})$, we compute
\[
\mathbf{J}^{\text{smpl}} 
= \mathcal{F}(\mathbf{R}, \mathbf{\Theta}),
\qquad
\hat{\mathbf{J}}^{\text{smpl}} 
= \mathcal{F}(\hat{\mathbf{R}}, \hat{\mathbf{\Theta}}).
\]
We then penalize discrepancies between reconstructed and ground-truth SMPL joints:
\begin{equation}
\mathcal{L}_{\text{smpl}} 
= \|\hat{\mathbf{J}}^{\text{smpl}} - \mathbf{J}^{\text{smpl}}\|.
\end{equation}
In addition, we enforce consistency between the reconstructed joint features $\hat{\mathbf{J}}$ in the gesture representation and the joints obtained by forwarding reconstructed rotations through SMPL:
\begin{equation}
\mathcal{L}_{\text{cons}} 
= \|\hat{\mathbf{J}} - \hat{\mathbf{J}}^{\text{smpl}}\|.
\end{equation}

To reduce training overhead introduced by SMPL-X forward passes, the 3D joint loss is computed on sparsely sampled frames with a stride of 4 starting from epoch 40. The loss weights are set to $\lambda_{\text{rec}}=1$, $\lambda_{\text{kl}}=10^{-6}$, $\lambda_{\text{smpl}}=10$, $\lambda_{\text{cons}}=10$, $\lambda_{\Delta j}=100$, and $\lambda_{\Delta r}=100$, selected via hyperparameter tuning on a validation set.

\revise{\noindent\textbf{Causal Receptive Field.}
In Eq.~(2), we use $g_{\leq i}$ as a shorthand for the causal input support available to latent token $z_i$. Since $i$ indexes latent tokens rather than motion frames, the actual support is determined by the bounded receptive field of the causal convolutional encoder. For a stack of causal convolutional layers indexed by $l$, with kernel size $k_l$, stride $s_l$, dilation $d_l$, and cumulative stride $J_l=\prod_{m<l}s_m$, the temporal receptive field is
\[
R = 1 + \sum_{l} (k_l - 1) \cdot d_l \cdot J_l
\]
In our implementation, the encoder has a total temporal downsampling factor of $s=4$ and a receptive field of $R=176$ frames. 
Therefore, for a zero-indexed latent token $z_i$, the full causal receptive field before clipping is
\[
\mathcal{R}_i =
\left[
\max(0, 4i - 172),\; 4i + 3
\right].
\]
Here, $4i+3$ is the last motion frame aligned with latent token $z_i$, and $172=R-4$ accounts for the receptive field extending backward from the aligned four-frame block. Since each primitive contains only $F_p=48$ frames and $R>F_p$, the receptive field is clipped at the start of the primitive. Thus, within a primitive, the effective support simplifies to
\[
\mathcal{R}_i = [0,\;4i+3].
\]
For example, $z_0$ is computed from frames $0$--$3$, $z_1$ from frames $0$--$7$, and the last latent token $z_{11}$ from frames $0$--$47$. In this sense, $q_\phi(z_i \mid g_{\leq i})$ conditions only on the causal prefix available to $z_i$ within the gesture primitive. Importantly, the encoder never accesses frames beyond the temporal position of $z_i$. This factorization is exact for the implemented causal encoder architecture; only the VAE posterior distribution itself remains variational.}

\subsection{Implementation details.}
The CausalVAE encoder consists of stacked causal 1D convolutional blocks with temporal downsampling. Two downsampling blocks with stride $2$ reduce the primitive length by a factor of $s=4$. For primitives of length $F_p=48$, this produces a latent sequence of length $l=12$. Each latent token has dimension $d=256$.
The encoder first projects the input motion features ($D=666$) to a hidden width of $1024$ using a causal convolution. Each downsampling stage applies a strided causal convolution followed by a causal residual stack with $3$ layers and exponentially increasing dilation ($1,3,9$). The final encoder layer maps the hidden representation to Gaussian posterior parameters $(\boldsymbol{\mu}_i,\boldsymbol{\sigma}_i^2)$ through a linear projection. Latent variables are sampled using the standard reparameterization trick.
The decoder mirrors the encoder architecture. Latent tokens are first projected to width $1024$, then processed by causal residual blocks and temporally upsampled, followed by causal convolutions to reconstruct the motion feature sequence.


\section{Latent Denoiser Model}\label{app:latent_denoiser}

\subsection{Conditioning signals}
\noindent\textbf{Audio features.}
We encode speech audio with a frozen HuBERT encoder~\cite{hsu2021hubert} and concatenate low-level prosody features, yielding
$\mathbf{a}\in\mathbb{R}^{F_a \times D_a},$
where $D_a=768$ for HuBERT base, and $D_a=768+D_p$ when concatenated with $D_p=7$ prosody features that are extracted per frame, consisting of $\log f_0$ (log fundamental frequency estimated via pitch tracking), energy (frame-level RMS amplitude), $\Delta \log f_0$ and $\Delta$energy (first-order temporal differences), harmonicity (energy normalized by a local energy window as a proxy for voicing strength), speech rate (local average of $|\Delta$energy$|$), and a pause indicator (binary flag for low-energy frames).

\noindent\textbf{Text features and spans.}
We extract CLIP token embeddings $\mathbf{t}\in\mathbb{R}^{F_t \times D_t}$ with a frozen CLIP text encoder ($D_t=512$)~\cite{radford2021learning}. In addition, each token is associated with an approximate span obtained by WhisperX~\cite{bain2022whisperx} in motion-frame indices, denoted
\begin{equation}
\mathbf{s}\in\mathbb{R}^{F_t \times 2}, \qquad \mathbf{s}_j = (u_j, v_j),
\end{equation}
where $(u_j, v_j)$ specifies the start and end frames of the token in the underlying motion timeline. These spans are used to form cross-attention masks.

\noindent\textbf{Posture Encoder Details.}
The initial posture reference $\gamma$ consists of SMPL-X joints $0$--$11$ from the first motion frame. For each joint $j$, we concatenate its 6D rotation and 3D position:
$
\mathbf{u}_j =
[\boldsymbol{\theta}^{6D}_j;\mathbf{q}_j]
\in\mathbb{R}^{9}.
$
A shared MLP projects each $\mathbf{u}_j$ to the transformer embedding dimension, and a learned joint-ID embedding is added to distinguish the corresponding body joints. Together with the learned sitting/standing token, this yields 13 posture tokens, $\mathbf{P}\in\mathbb{R}^{13\times d_e}$. 

\noindent\textbf{Posture Modulation Baseline.}
For the modulation baseline in Tab.~\ref{tab:test_sets}, posture is represented by a binary sitting/standing label. We map this label to a learned embedding and add it to the diffusion timestep embedding before conditioning the denoiser tokens. This provides a coarse global posture cue but does not encode the detailed body configuration captured by our proposed initial-posture reference.

\subsection{Temporal Cross-Attention Masking}
\label{app:temporal_masks}

We support a masking mode for audio cross-attention that gives flexibility and alignment strength.
Each latent index attends to a local window around its aligned audio index. This imposes a soft monotonic alignment and improves word-level synchronization when audio is streamed.
We also restrict latent indices that fall inside the span of a text token to attend to the matching text token. This yields an explicit alignment controller without changing the denoiser architecture.

Formally, let $i \in \{1,\dots,l\}$ index future latent tokens and 
$j \in \{1,\dots,F_a+F_t\}$ index conditioning tokens (audio and text). 
We define a binary cross-attention mask $\mathbf{M} \in \mathbb{R}^{l \times (F_a+F_t)}$ that is added to the attention logits before softmax:
\begin{equation}
\mathrm{Attn}(\mathbf{Q},\mathbf{K},\mathbf{V})
=
\mathrm{softmax}\!\left(
\frac{\mathbf{Q}\mathbf{K}^\top}{\sqrt{d_e}}
+
\mathbf{M}
\right)\mathbf{V}.
\end{equation}
Entries of $\mathbf{M}$ are defined as
\begin{equation}
\mathbf{M}_{i,j}
=
\begin{cases}
0, & \text{if } j \in \mathcal{W}(i), \\
-\infty, & \text{otherwise},
\end{cases}
\end{equation}
where $\mathcal{W}(i)$ is the allowed conditioning window for latent index $i$.

\vspace{5pt}
\noindent\textbf{Audio Window Attention.}
Let $\phi(i)$ denote the temporally aligned audio index for latent token $i$. 
We define a local window of radius $r$:
\begin{equation}
\mathcal{W}_a(i)
=
\left\{
j \mid |\; j - \phi(i)\; | \le r
\right\}.
\end{equation}
This enforces soft monotonic alignment by restricting each latent token to attend only to nearby audio frames, improving rhythmic consistency under streaming speech.

\vspace{5pt}
\noindent\textbf{Text Span Attention.}
Let $\mathcal{T}_j = [s_j, e_j]$ denote the temporal span of text token $j$ in the audio, where $s_j$ and $e_j$ are its start and end times. For each latent index $i$, we restrict attention to the text tokens whose temporal spans include $i$:
\begin{equation}
\mathcal{W}_t(i)
=
\{\, j \mid \psi(i) \in [s_j, e_j] \,\}.
\end{equation}
where $\psi(i)$ maps gesture latent index $i$ to its corresponding xsequence frame.
Thus, latent generation at gesture latent index $i$ is conditioned only on the text tokens aligned with that time.
The final conditioning window is:
\begin{equation}
\mathcal{W}(i)
=
\mathcal{W}_a(i)
\cup\
\mathcal{W}_t(i).
\quad
\end{equation}

This masking strategy enables explicit temporal control without modifying the transformer architecture.

\subsection{Implementation Details}

The latent denoiser is a $7$-layer transformer with embedding dimension $512$, $8$ attention heads, feed-forward dimension $2048$, and dropout $0.1$. Motion history and noisy latent tokens are projected to the shared embedding space and combined with positional and timestep embeddings. Conditioning uses frozen HuBERT+prosody audio features, frozen CLIP text embeddings, posture tokens, and a learnable speaker embedding added to the timestep embedding. Each transformer layer contains an independent zero-initialized posture-attention gate.

Training uses classifier-free guidance with dropout with probability $0.2$, a DDPM scheduler with $1000$ timesteps, and a scaled linear noise schedule. Inference uses DDIM with $20$ steps and guidance scale $2.5$. Unless otherwise stated, audio cross-attention uses window masking with $\xi=1.0$.

\section{Stage3 Details}
\label{app:stage3}
\subsection{Collision Loss}
To obtain a GPU-efficient collision loss, we adapt the implementation of MOVER \cite{yi2022human} by constructing a signed distance field (SDF) grid of size $32^3$ for each object. In this volumetric grid, each voxel stores the minimum signed distance from its center to the object mesh. As in standard SDF representations, distances are positive outside the mesh and negative inside.

In MOVER, the sign is determined by casting a ray from the voxel in a certain predefined direction: if the ray intersects the mesh an odd number of times, the voxel is classified as inside; if the number of intersections is even, it is classified as outside. While this works in theory, the approach can be unstable in practice because the ray direction may align with the surface, leading to unpredictable results. To improve robustness, we cast rays in 50 random directions and use majority voting to determine the sign. 

Given that object scales can vary significantly (e.g., a couch versus a chair), we construct the grid only within the BPS ellipsoid. During training, we extract the SMPL-X vertices from the generated motion and filter out the lower body. For each upper-body vertex $u_i$ at frame $i$, we query $f(u_i)$ to determine the corresponding voxel and compute the collision loss as follows:
\begin{equation}
    \mathcal{L}_{\text{collision}} 
    = \sum_{i} \sum_{u} \left\| \hat{V}_{f(u_i)} \right\|_2^2,
    \quad \hat{V}_{f(u_i)} < 0 .
\end{equation}

\subsection{Implementation Details}

Stage 3 initializes from the pretrained Stage 2 model and freezes all parameters except the BPS projection and the gated object fusion modules. Object–human relations are encoded using a 1024-dimensional BPS descriptor and injected into each transformer block via the gated fusion module.


Training follows the same diffusion setup as Stage 2 while adding the collision loss. We train with AdamW using learning rate $10^{-3}$, batch size $128$, and loss weights $\lambda_{x_0}=1$, $\lambda_{\text{latent}}=1$, and $\lambda_{\text{collision}}=50$.

\begin{figure*}[t]
    \centering
    \includegraphics[width=0.8\linewidth]{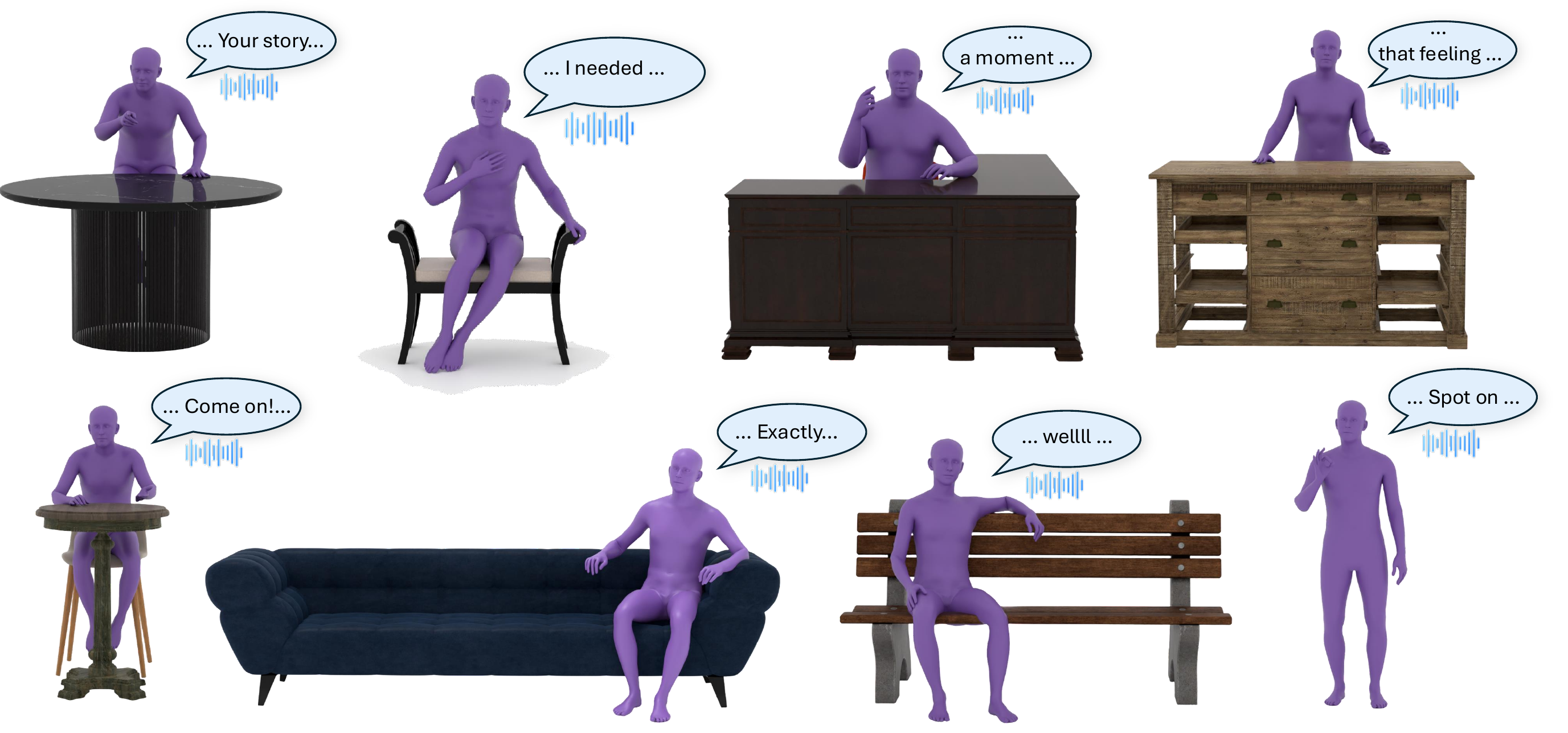}
    \caption{\footnotesize Sample frames from our \sceneges{} dataset showing diverse conversational gestures performed in different scene contexts (tables, chairs, desks, couches, benches). The dataset captures natural speech-driven gestures with varying postures. Each sequence includes synchronized speech, full-body motion, and scene geometry. Video examples are shown in the supplementary video.}
    \label{fig:app_db}
\end{figure*}

\section{\sceneges{} Dataset} \vspace{-5pt} \label{app:db}

Existing gesture datasets focus on speech-driven motions captured in standing, monocular, or controlled studio settings. Although they provide high-quality motion data, they lack explicit modeling of surrounding objects. In reality, gestures are shaped by the environment, as people rest their hands on tables, lean on armrests, avoid obstacles, and adjust movements to nearby furniture. Without object representations, models cannot learn object-grounded gestural behavior. To address this limitation, we introduce \sceneges, a synthetic dataset for scene-aware co-speech gesture generation. It includes diverse chairs and tables with varying geometries and topologies paired with 3D SMPL-X~\cite{pavlakos2019expressive} gesture motions, enabling object-grounded gesture learning.

\subsection{Object Collection and Scene Construction.} We curate a diverse set of 3D objects from the HSSD~\cite{khanna2024habitat} dataset, including chairs and tables that span a wide range of topologies. This diversity introduces variation in support surfaces, armrests, heights, and overall spatial affordances. To construct the scene, we define three simple layouts and place either a chair, a table/countertop, or both alongside a human with varied initial poses and appearances. 

\subsection{Scenario and Dialogue Generation} To construct semantically coherent and object-aware conversational data, we adopt a staged generation process driven by the Gemini~\cite{team2023gemini} model. We first generate high-level interaction scenarios (see.~\cref{fig:scenario_prompt}) that define the conversational setting (e.g., formal discussion, casual meeting, etc.). Given each scenario, Gemini produces multi-turn dialogue scripts (see.~\cref{fig:dialogue_prompt}) and, importantly, explicitly specifies the intended human gestures accompanying each utterance. These gesture descriptions include grounded behaviors such as resting hands on the table, leaning toward the surface while emphasizing a point, or constraining arm movement near armrests (see.~\cref{fig:gesture_prompt}).
Separately, for each selected chair–table asset pair, we render a frontal-view image of the instantiated scene. This rendered first frame is used as visual conditioning for Veo~3~\cite{deepmind2024veo3}. The scenario, dialogue, and gesture specifications are then converted into a structured video prompt (\cref{fig:video_prompt}), and Veo~3 generates the complete motion sequence starting from the initialized frame, keeping the object geometry fixed while animating the human performer.

\subsection{Gesture Extraction and Refinement} After video synthesis, we recover 3D human motion using SAM 3D Body~\cite{yang2026sam} to obtain per-frame body representations. Because generative video may introduce minor spatial inconsistencies, the extracted motions are manually refined by a professional artist. The refinement is performed frame-by-frame to ensure that the reconstructed motion faithfully follows the video while removing object penetrations and correcting contact states with chairs and tables. This step results in physically plausible, collision-free, and scene-consistent gesture sequences suitable for training grounded models.

\subsection{Dataset Statistics.}
\sceneges{} contains 26 interaction scenarios instantiated across $153$ object assets ($106$ chairs and $47$ tables). Each video has an average duration of $8$ seconds, yielding a total of $28$ minutes of object-grounded co-speech motion. The dataset covers diverse conversational contexts, object topologies, and contact configurations between humans and surrounding furniture.

\revise{\subsection{Dataset Validation and Perceptual Realism}
Because \sceneges{} is synthetically generated and then refined, we further validate whether its motions are perceptually and kinematically comparable to real conversational motion. We conduct two complementary analyses: a user study comparing \sceneges{} with real motion from Embody3D, and a kinematic validation comparing the distribution of motion statistics between the two datasets.

\subsubsection{\sceneges{} Human Perception Study}
We evaluated perceptual realism through a two-alternative forced-choice (2AFC) and Likert-scale evaluation with 23 participants.
Participants were asked to evaluate 65 randomly sampled motion clips from \sceneges{} and Embody3D. For each clip, participants answered whether the motion appeared real or synthetic and rated its naturalness on a 0 to 5 scale. The real/synthetic classification results are shown in Fig.~\ref{fig:user_study_confusion}. The balanced accuracy is 51.41\%, close to random chance, indicating that participants could not reliably distinguish \sceneges{} from real captured motion. Furthermore, \sceneges{} achieved an average naturalness score of 4.28, performing competitively against—and even slightly exceeding—the real Embody3D average of 4.09.
These results suggest that \sceneges{} achieves perceptual naturalness comparable to real conversational motion, despite being generated through a synthetic data pipeline.

\begin{figure}
    \centering
    \includegraphics[width=0.55\linewidth]{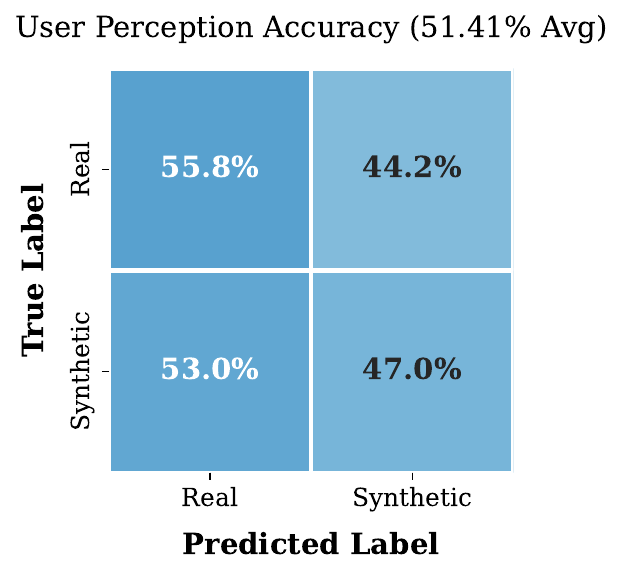}
    \caption{\revise{User Study Confusion Matrix.}}  \vspace{-10pt}
    \label{fig:user_study_confusion}
    \vspace{-15pt}
\end{figure}

\subsubsection{Quantitative kinematic validation}
To complement the perceptual study, we compare \sceneges{} with Embody3D using kinematic motion statistics computed from upper-body gestures. As shown in Fig.~\ref{fig:sceneges_kinematic_validation}, \sceneges{} closely matches the central tendency and spread of real conversational motion across gesture-related metrics. This indicates that the generated and artist-refined motions do not only appear natural to human observers, but also follow motion distributions similar to real conversational gestures.

\begin{figure}
    \centering
    \includegraphics[width=1\linewidth]{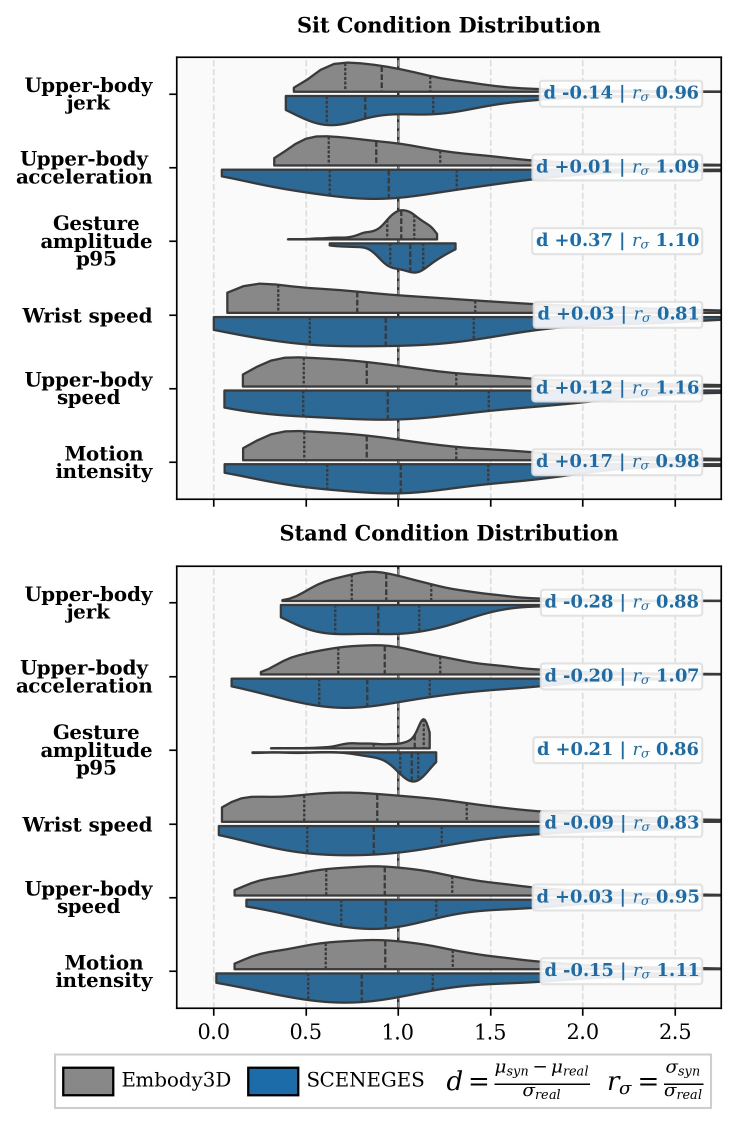}
    \caption{\revise{Kinematic validation of \sceneges{} against Embody3D. We compare distributions of upper-body gesture statistics between \sceneges{} and real conversational motion. }}
    \label{fig:sceneges_kinematic_validation}
\end{figure}}
\section{More Experiments and Ablations}



\begin{table}[t]
\centering
\small
\setlength{\tabcolsep}{6pt}
\caption{Effect of latent dimensionality on reconstruction performance evaluated on BEAT2 and Embody3D. Lower is better for all metrics.}
\begin{tabular}{c l c c c}
\toprule
Dataset & Dimension & FGD$\downarrow$ & MPJPE$\downarrow$ & ACCL$\downarrow$ \\
\midrule

\multirow{4}{*}{BEAT2}
& 6$\times$256  & 1.531 & 11.983 & 7.155 \\
& 6$\times$512  & \underline{0.127} & 8.384 & 7.063 \\
& 12$\times$256 & 0.168 & \underline{7.594} & \textbf{6.624} \\
& 12$\times$512 & \textbf{0.048} & \textbf{6.813} & 6.747 \\
\midrule

\multirow{4}{*}{Embody3D}
& 6$\times$256  & 0.912 & 14.685 & 6.045 \\
& 6$\times$512  & \underline{0.089} & 10.96  & 5.753 \\
& 12$\times$256 & 0.153 & \underline{9.401}  & \textbf{5.507} \\
& 12$\times$512 & \textbf{0.037} & \textbf{8.225} & 5.698 \\
\bottomrule
\end{tabular}
\label{tab:latent-dim-causalvae-recon}
\end{table}


\begin{table}[t]
\centering
\small
\setlength{\tabcolsep}{6pt}
\caption{Effect of latent dimensionality on generation performance on BEAT2.}
\vspace{-10pt}
\begin{tabular}{c l c c c c}
\toprule
Dataset & Dimension & FGD$\downarrow$ & Div$\uparrow$ & BC$\uparrow$ & $\Delta$BC$\downarrow$ \\
\midrule

\multirow{4}{*}{BEAT2}
& 6$\times$256  & 4.262 & 10.091 & 7.429 & 0.234 \\
& 6$\times$512  & 3.709 & 10.908 & 7.542 & 0.379 \\
& 12$\times$256 & \textbf{3.436} & \textbf{14.131} & \textbf{7.693} & \textbf{0.201} \\
& 12$\times$512 & 3.7.95 & 11.034 & 7.352 & 0.397 \\

\bottomrule
\end{tabular}\vspace{-10pt}
\label{tab:latent-dim-generation}
\end{table}

We present additional ablation studies to further analyze key components of the proposed framework. Specifically, we study the effect of the latent dimensionality in CausalVAE, the influence of the reweighting coefficient $\lambda_I$, the role of classifier-free guidance during generation, and the impact of the number of diffusion sampling timesteps. These experiments provide additional insight into the design choices and performance trade-offs of the model.

\subsection{\puppet{} Human Perception Study}\vspace{-3pt}

\subsection{Effect of CausalVAE Latent Size}\vspace{-3pt}

We investigate the effect of the latent dimensionality of CausalVAE on both reconstruction quality and downstream gesture generation. \cref{tab:latent-dim-causalvae-recon} reports reconstruction performance, while \cref{tab:latent-dim-generation} evaluates generation performance.
Increasing the latent capacity generally improves reconstruction accuracy. Configurations with larger latent representations (e.g., $12\times512$) achieve the lowest FGD and MPJPE on both datasets, indicating that higher capacity allows the encoder to preserve more motion details. However, this improvement in reconstruction does not directly translate to better generative performance. As shown in \cref{tab:latent-dim-generation}, overly large latent spaces lead to reduced diversity and weaker speech–gesture synchronization.

Among all configurations, the $12\times256$ latent size achieves the best overall generation quality, yielding the lowest FGD and $\Delta$BC together with the highest BC and diversity. This suggests that increasing the number of latent tokens improves temporal expressiveness, while keeping the token dimensionality moderate maintains a compact and well-structured latent space that is easier for the diffusion model to learn. In contrast, larger token dimensions (e.g., $12\times512$) increase the latent complexity without improving generation metrics.

Based on these observations, we adopt the $12\times256$ latent configuration in all experiments, as it provides the best balance between reconstruction fidelity, generation quality, and latent compactness.

\begin{table}[h]
\centering
\small
\setlength{\tabcolsep}{6pt}
\caption{Effect of classifier-free guidance (CFG) scale on generation performance.}
\vspace{-8pt}
\begin{tabular}{c c c c c}
\toprule
CFG Scale & FGD$\downarrow$ & Div$\uparrow$ & BC$\uparrow$ & $\Delta$BC$\downarrow$ \\
\midrule
1.0 & 7.093 & 14.063 & 7.394 & 0.207 \\
1.5 & 7.509 & 14.108 & 7.509 & 0.262 \\
2.0 & 3.648 & 14.118 & 7.637 & 0.355 \\
2.5 & \textbf{3.436} & \textbf{14.131} & 7.693 & \textbf{0.201} \\
3.0 & 3.738  & 13.726 & 7.723 & 0.427 \\
4.0 & 4.917 & 13.970 & 7.772 & 0.479 \\
5.0 & 6.774 & 13.653 & 7.786 & 0.489 \\
7.0 & 11.775 & 14.006 & \textbf{7.891} & 0.596 \\
\bottomrule
\end{tabular}\vspace{-10pt}
\label{tab:cfg-ablation}
\end{table}

\subsection{Effect of Classifier-Free Guidance Scale}
We study the effect of the CFG scale on gesture generation performance in \cref{tab:cfg-ablation}. When the guidance scale is low (e.g., 1.0–1.5), the model produces diverse gestures but with relatively poor realism, as reflected by higher FGD scores and weaker speech–gesture synchronization.

Increasing the CFG scale improves alignment with the conditioning signals, leading to a substantial improvement in FGD and BC. The best overall performance is achieved at a CFG scale of $2.5$, which yields the lowest FGD and $\Delta$BC while maintaining the highest diversity. This indicates a good balance between generation quality and adherence to the speech conditioning.

However, further increasing the guidance strength degrades performance. Large CFG values overly constrain the generation process, which leads to higher FGD and reduced diversity, while also increasing the synchronization gap ($\Delta$BC). Based on these results, we use a CFG scale of $2.5$ in all experiments, as it provides the best trade-off between realism, diversity, and speech–gesture synchronization.

\begin{table}[t]
\centering
\small
\setlength{\tabcolsep}{6pt}
\caption{Effect of number of diffusion sampling steps on generation performance.}
\vspace{-8pt}
\begin{tabular}{c c c c c}
\toprule
Steps & FGD$\downarrow$ & Div$\uparrow$ & BC$\uparrow$ & $\Delta$BC$\downarrow$ \\
\midrule
1   & 4.930 & 10.911 & 7.692 & 0.402 \\
10  & 3.441 & 13.698 & 7.682 & 0.387 \\
20  & \textbf{3.436} & \textbf{14.131} & \textbf{7.693} & \textbf{0.201}   \\
50  & 3.448 & 14.057 & 7.680 & 0.382 \\
100 & 3.455 & 14.056 & 7.674 & 0.380 \\
\bottomrule
\end{tabular}\vspace{-8pt}
\label{tab:steps-ablation}
\end{table}

\subsection{Effect of number of sampling timesteps}

As shown in \cref{tab:steps-ablation}, even with a single sampling step the model produces competitive results compared to prior methods (Tab.~2), indicating that the learned latent representation provides a strong initialization for generation. However, using very few steps limits the refinement of the latent motion, leading to higher FGD and lower diversity.
Increasing the number of sampling steps further improves performance. Moving from 1 to 10 steps substantially reduces FGD and increases diversity, suggesting that additional denoising iterations allow the model to better refine motion details and speech–gesture alignment. The best overall performance is achieved at 20 sampling steps, which yields the lowest FGD and $\Delta$BC together with the highest diversity and strong BC.
Further increasing the number of sampling steps does not improve performance. 
Based on these results, we use 20 sampling steps in all experiments as it provides the best trade-off between generation quality and sampling efficiency.

\begin{table}[t]
\centering
\scriptsize
\setlength{\tabcolsep}{9pt}
\caption{Ablation on the intensity reweighting coefficient $\lambda_I$ on BEAT2 (Speaker2) and the full dataset.}
\vspace{-10pt}
\begin{tabular}{l c c c c c}
\toprule
Dataset & $\lambda_I$ & FGD$\downarrow$ & Div$\uparrow$ & BC$\uparrow$ & $\Delta$BC$\downarrow$\\
\midrule

\multirow{4}{*}{BEAT2 (Speaker2)}
 & 0 & \textbf{2.887} & 11.083 & 7.566 & 0.313 \\
 & 1 & 3.167 & 11.642 & 7.608 & 0.335 \\
 & 2 & 3.484 & 12.802 & 7.601 & 0.317 \\
 & 3 & 3.437 & \textbf{14.131} & \textbf{7.693} & \textbf{0.201} \\
 & 4 & 3.544 & 14.001 & 7.678 & 0.231 \\
\midrule

\multirow{4}{*}{BEAT2 (All)}
 & 0 & 3.887 & 9.881 & 6.658 & 0.627 \\
 & 1 & 5.221 & 9.783 & 6.666 & 0.624 \\
 & 2 & 4.157 & 11.481 & 6.759 & 0.561 \\
 & 3 & \textbf{3.550} & 12.062 & 6.791 & \textbf{0.431} \\
 & 4 & 3.901 & \textbf{12.117} & \textbf{6.806} & 0.445 \\
\bottomrule
\end{tabular}
\label{tab:lambda}
\end{table}

\begin{table}[t]
\setlength{\tabcolsep}{5pt}
    \centering \footnotesize
    \caption{\footnotesize Baseline comparison for co-speech gesture generation on the BEATv2 (Speaker2) test set. All metrics are reported in  $\times 10^{-1}$.}
    \vspace{-0.1in}
    \begin{tabular}{lccccc}
        \toprule
        Methods & Train & Test
        & FGD$\downarrow$ & BC$\uparrow$ & $\Delta$BC$\downarrow$\\
        \midrule
        EMAGE~\cite{liu24emage}            & Speaker2 & Speaker2 & 5.117 & 5.910 & 1.522 \\
        LOM~\cite{chen2025language}         & Speaker2 & Speaker2 & 4.538 & 6.114 & 1.318 \\
        GestureLSM~\cite{liu2025gesturelsm} & Speaker2 & Speaker2 & 3.692 & 7.547  & \textbf{0.166}  \\
        \midrule
        \puppet{} (Ours)                     & Speaker2 & Speaker2 & \textbf{3.436} & \textbf{7.693} &  0.201 \\
        \puppet{} (Ours)                   &  BEAT2  & Speaker2    & 4.663 & 7.347 & 0.228 \\
        \bottomrule
    \end{tabular}
    \vspace{-0.1in}
    \label{tab:sota_compare2}
\end{table}


\subsection{Effect of coefficient $\lambda_I$}
\cref{tab:lambda} investigates the effect of the reweighting coefficient $\lambda_I$ (Eq.~7), which controls the strength of intensity-based loss weighting during training. Since Speaker2 is a highly expressive actor with strong gesture intensity, we additionally evaluate the effect of $\lambda_I$ on the full dataset to verify that the observations generalize beyond this speaker.

When $\lambda_I=0$, no reweighting is applied and the model achieves the lowest FGD for the single-speaker setting, but produces lower diversity and weaker speech–gesture synchronization. Increasing $\lambda_I$ encourages the model to emphasize motion segments with higher gesture intensity, which improves diversity and speech–gesture alignment. As $\lambda_I$ increases, both BC and Div generally improve while $\Delta$BC decreases, indicating stronger synchronization between speech and generated gestures.

The best overall trade-off is achieved at $\lambda_I=3$, which yields the highest BC and diversity together with the lowest $\Delta$BC for the single-speaker setting, and the best synchronization performance on the full dataset. Larger values provide no consistent improvement and slightly degrade some metrics. Based on these results, we set $\lambda_I=3$ in all experiments.

\subsection{Single-Speaker vs Multi-Speaker Training}

Previous works typically train and evaluate their models on a single speaker (e.g., Speaker2 in BEAT2), as prior studies report that performance, particularly FGD, degrades when training on multiple speakers~\cite{liu24emage}. Consequently, many methods report results only under the single-speaker setting.
Table~\ref{tab:sota_compare2} compares this setting with our model trained either on Speaker2 only or on the full BEAT2 dataset. When trained and evaluated on Speaker2, \puppet{} achieves the best FGD and BC among all methods while maintaining competitive $\Delta$BC. Importantly, when trained on the full multi-speaker BEAT2 dataset and evaluated on Speaker2, the model still achieves strong performance, demonstrating that \puppet{} generalizes well across speakers.

\revise{\subsection{Generalization to Unseen Objects (Held-Out Split).} \label{app:unseen_obj}
To explicitly verify that \puppet{} learns generalized, object-conditioned gesture priors rather than merely memorizing localized 3D scene geometry, we establish a zero-shot evaluation on a completely held-out object asset split. During testing, the model is evaluated in novel structural configurations containing geometric assets entirely unseen during training. 


Despite the lack of exposure to these specific object boundaries, \puppet{} gracefully preserves its spatial precision, yielding a \textit{MeanPen} of $3.280 \times 10^{-4}$ and an \textit{LL1} error of $0.86$. These metrics closely align with our main results in \cref{tab:abl-collision}, demonstrating the model's capacity to lift spatial affordances dynamically from structural context and synthesize robust, posture-aware movements across arbitrary unseen furniture.
}


\begin{figure*}[t]
    \centering
    \includegraphics[width=0.85\linewidth]{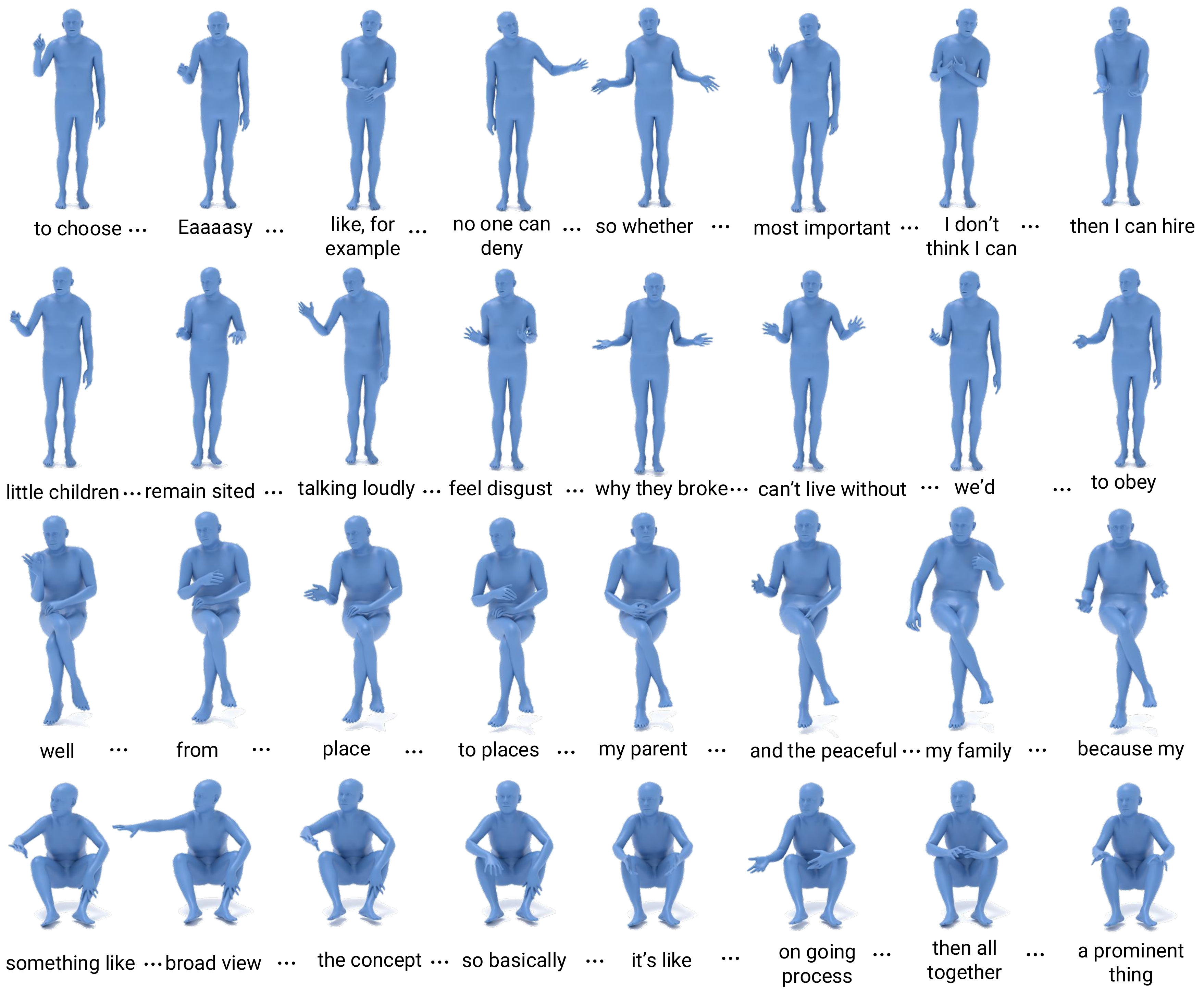}
    \caption{Additional qualitative results. The examples demonstrate diverse co-speech gestures across both standing and sitting postures while maintaining temporal alignment with speech.} \vspace{-10pt}
    \label{fig:more_qualitative}
\end{figure*}
\section{More Qualitative Results}
\label{app:qualitative}
Fig.~\ref{fig:more_qualitative} presents additional qualitative examples generated by \puppet{} across different speech audios and postures. The results illustrate that the model produces diverse gestures that remain temporally aligned with the spoken words while maintaining consistent posture. The examples include both standing and sitting scenarios, demonstrating that the model adapts gesture formation according to the underlying posture while preserving natural conversational motion patterns (see video for dynamic results).

\begin{figure*}
    \centering
    \includegraphics[width=0.85\linewidth]{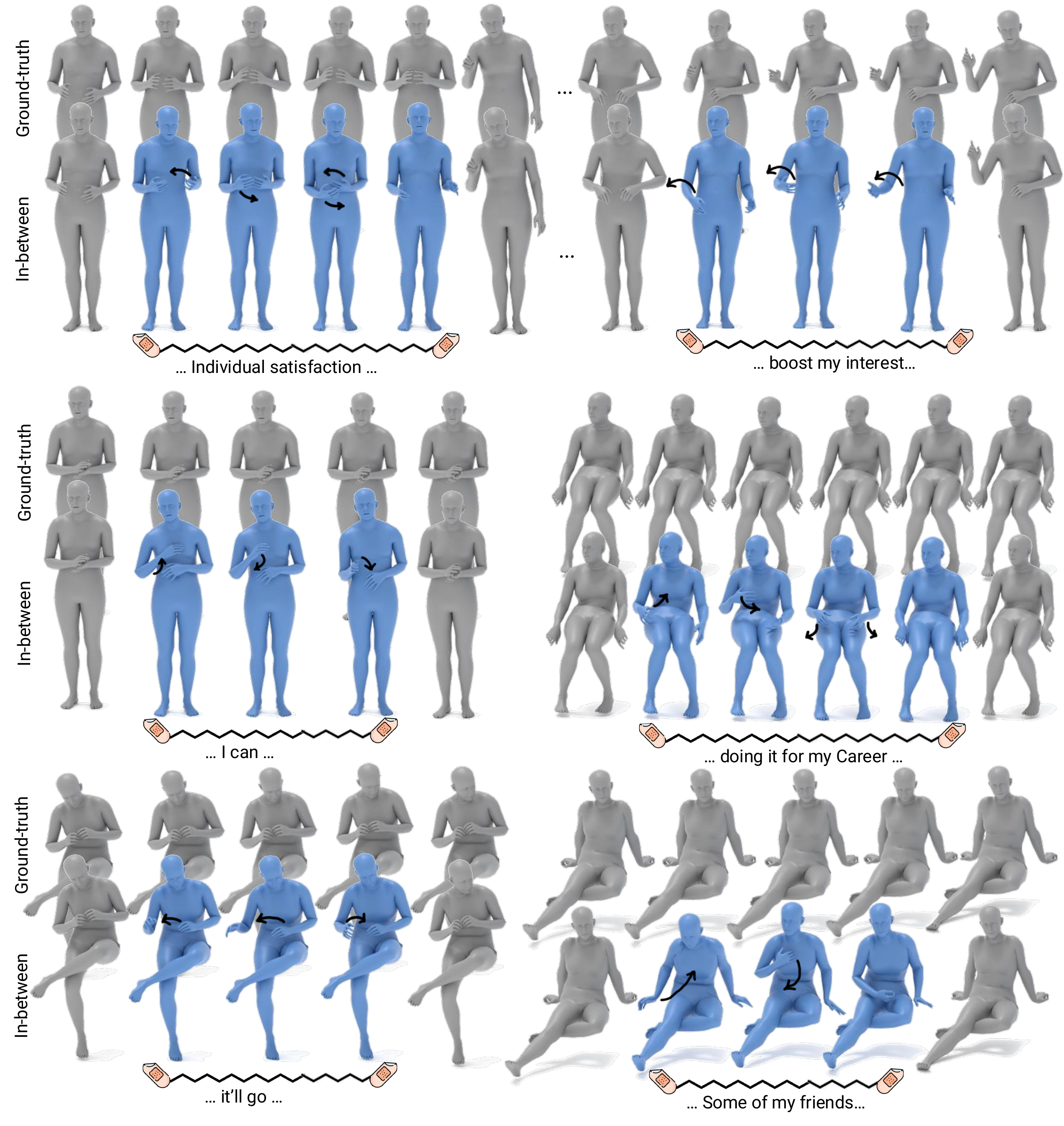}
    \caption{Gesture in-betweening and completion examples. Gray: ground truth, blue: regenerated intermediate frames.}
    \label{fig:inbetween_completion}
\end{figure*}
\subsection{Gesture In-betweening and Completion}

Because \puppet\ operates in a causal latent gesture space, it naturally supports gesture in-betweening and completion. 
\cref{fig:inbetween_completion} shows examples where a middle segment of motion is regenerated between surrounding ground-truth frames. 
The generated poses (blue) form smooth transitions while producing gestures that better match the spoken content. 
In several cases, the regenerated segment replaces a less expressive motion with a more appropriate gesture, which can be useful when producing animations that better match the intended expression.
Additional qualitative examples for this task are provided in the supplementary video.

\subsection{Effect of Posture Conditioning}

\begin{figure}
    \centering
    \includegraphics[width=1\linewidth]{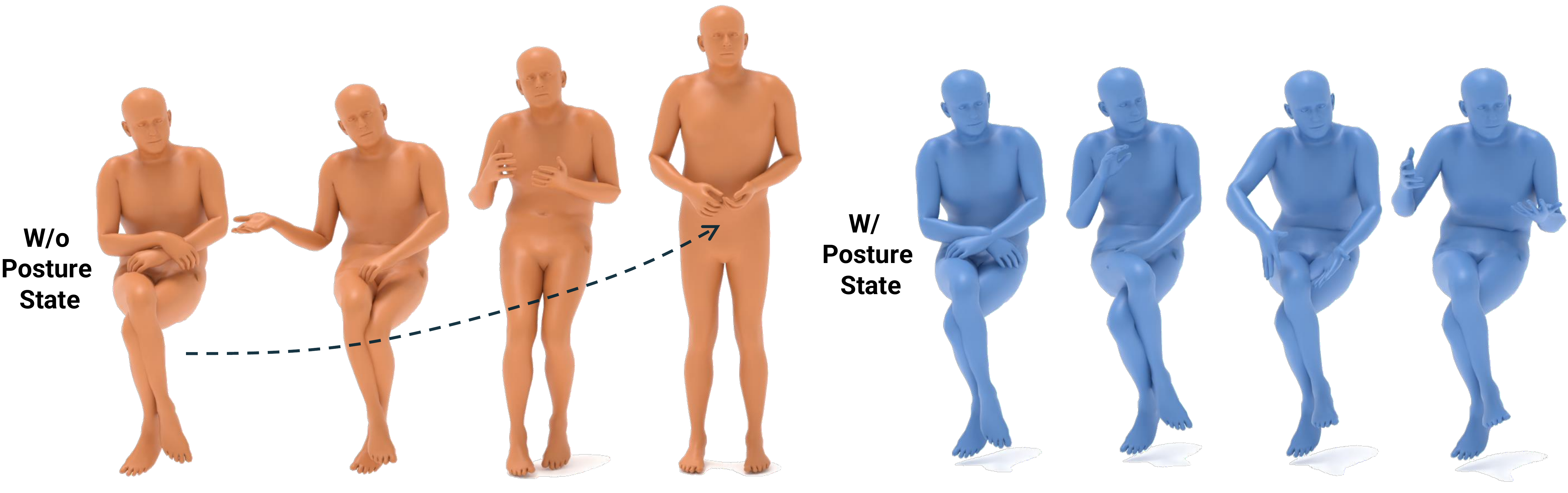}
    \caption{\small Effect of posture conditioning. Without explicit posture conditioning, the generated motion gradually drifts from the intended body configuration, while the initial-posture reference helps preserve it throughout generation.}
    \label{fig:posturestate}
\end{figure}

We further examine the effect of conditioning the model on the posture state. 
\cref{fig:posturestate} compares results generated without and with the posture condition. 
Without this signal, the model tends to drift toward a neutral standing pose during generation, as standing poses are more frequently observed during training. 
In contrast, when the posture condition is provided, the model preserves the intended body configuration (e.g., sitting) while generating appropriate gestures.

\section{More on Related Work}
\noindent\textbf{Speech-driven gesture generation.}
Speech-driven gesture generation aims to synthesize human motion that is temporally aligned with speech and semantically meaningful, leveraging signals such as audio, language, and speaker characteristics~\cite{nyatsanga2023review,ginosar2019learning,yoon2020trimodal,ahuja2020nogestures,kucherenko2020gesticulator}. 

Early work focuses on designing architectures that effectively fuse multimodal cues to produce natural and expressive gestures~\cite{ferstl2018investigating,ginosar2019learning,yoon2020trimodal,kucherenko2020gesticulator}. CaMN~\cite{liu22beat} proposes a cascaded multimodal fusion framework that progressively integrates audio, text, emotion, and speaker identity signals, enabling conversational gesture synthesis with richer conditioning and improved expressiveness compared to purely audio-driven approaches. DisCo~\cite{liu2022disco} subsequently explores gesture diversity by disentangling rhythmic structure from semantic content, allowing the model to generate multiple plausible gesture realizations while maintaining synchronization with speech rhythm. EMAGE~\cite{liu24emage} advances the field toward holistic motion generation by jointly modeling face, hands, and body through masked modeling and large-scale pretraining, improving realism and cross-modal coherence across modalities. The Language of Motion~\cite{chen2025language} reframes gesture generation as a multimodal language modeling problem, unifying speech, text, and motion tokens within a shared representation space to improve generalization and flexibility across tasks. SemGes~\cite{liu2025semges} further emphasizes semantic grounding by combining audio and textual representations with a learned motion prior, encouraging gestures that reflect discourse-level meaning beyond simple beat alignment. ViBES~\cite{zhang2025vibes} extends gesture synthesis to conversational settings by jointly modeling speech, language, and behavior, enabling generation that accounts for interaction dynamics across turns and more complex conversational contexts.

Alongside these approaches, diffusion models represent an alternative paradigm for gesture synthesis, focusing on iterative refinement of motion through stochastic denoising processes~\cite{mughal2024convofusion, zhu2023taming}. DiffuseStyleGesture~\cite{yang2023diffusestylegesture} introduces diffusion-based generation with cross-local attention to better align speech and motion and enable stylistic control while operating directly in motion representations during denoising. SynTalker~\cite{chen2024enabling} proposes a diffusion framework with prompt-based control over different body regions, allowing fine-grained manipulation of generated gestures and extending beyond purely standing scenarios to include pose variations such as sitting. InteracTalker~\cite{rajan2025interactalker} introduces prompt-based human–object interaction control for co-speech gesture generation, enabling high-level specification of interaction behaviors through text prompts. Nevertheless, because it is pretrained on generic motion data and fine-tuned on gesture datasets without explicit modeling of scene context, the generated motions remain largely scene-agnostic, lacking alignment with surrounding objects and failing to produce physically grounded interactions such as adapting hand placement to nearby furniture. DiffSHEG~\cite{chen2024diffsheg} performs joint diffusion over facial expressions and body motion to generate synchronized full-body gestures with strong temporal coherence. GestureLSM~\cite{liu2025gesturelsm} explores efficient diffusion modeling through architectural shortcuts and structured token interactions that enable real-time gesture synthesis while maintaining expressiveness. GestureHYDRA~\cite{yang2025gesturehydra} combines diffusion with hybrid conditioning and retrieval mechanisms to activate semantically meaningful gestures, particularly emphasizing hand motion synchronized with speech cues. 

Despite substantial progress, existing methods often rely on local conditioning schemes where gesture tokens are combined with pooled audio representations over short temporal windows, which can limit the ability to capture longer-range dependencies and may discard fine-grained prosodic information~\cite{nyatsanga2023review}. While some approaches incorporate textual signals, they typically do not explicitly model the timing of words within speech, making it harder to fully exploit semantic cues for precise synchronization. To cope with limited motion diversity, many methods adopt discrete motion tokenization such as VQ-based representations~\cite{van2017neural}, which can introduce quantization artifacts and reduce fidelity in subtle motion dynamics~\cite{liu2022audiodriven,yang2023qpgesture,eggesture2024}. Furthermore, while recent work such as InteracTalker introduces prompt-level interaction control, these methods still lack explicit scene grounding, and therefore cannot ensure that generated gestures physically align with objects present in the environment.

\noindent\textbf{Scene-aware Motion Generation.}
Scene-aware motion generation focuses on synthesizing human behaviors that are physically consistent with surrounding environments and objects~\cite{hassan2019prox,hassan2021posa,li2019putting,zhang2020place,wang2021longterm,hassan2021samp}. DIMOS~\cite{zhao2023synthesizing} formulates human-scene interaction as a reinforcement learning problem, learning scene-conditioned control policies that enable virtual humans to navigate complex indoor environments and interact with objects such as sitting or lying while avoiding collisions. TeSMo~\cite{yi2024generating} introduces a text-conditioned diffusion framework that combines a scene-agnostic motion prior with a scene-aware control branch, allowing motions to be generated from textual descriptions while respecting obstacles and interaction targets through navigation and interaction modules. HSI-GPT~\cite{wang2025hsi} scales this direction by proposing a large scene-motion-language model that jointly reasons over scene context, human motion, and language to support diverse human-scene interaction tasks within a unified framework. Meanwhile, MOVER~\cite{yi2022human} leverages human-scene interaction cues such as contact, occlusion, and free-space constraints to jointly optimize scene layouts and human motion, improving physical consistency between humans and their environments. Despite these advances, existing works primarily focus on locomotion or task-oriented interactions, and there remains a lack of datasets that capture co-speech gestures naturally aligned with surrounding objects, limiting progress toward gesture generation that is both communicative and physically grounded in real scenes.

\noindent\textbf{Scene-aware Motion Generation.}
DIMOS~\cite{zhao2023synthesizing} formulates human-scene interaction as a reinforcement learning problem, learning scene-conditioned control policies that enable virtual humans to navigate complex indoor environments and interact with objects such as sitting or lying while avoiding collisions.
TeSMo~\cite{yi2024generating} introduces a text-conditioned diffusion framework that combines a scene-agnostic motion prior with a scene-aware control branch, allowing motions to be generated from textual descriptions while respecting obstacles and interaction targets through navigation and interaction modules.
HSI-GPT~\cite{wang2025hsi} scales this direction by proposing a large scene-motion-language model that jointly reasons over scene context, human motion, and language to support diverse human-scene interaction tasks within a unified framework.
Meanwhile, MOVER~\cite{yi2022human} leverages human-scene interaction cues such as contact, occlusion, and free-space constraints to jointly optimize scene layouts and human motion, improving physical consistency between humans and their environments.
Despite these advances, existing works primarily focus on locomotion or task-oriented interactions, and there remains a lack of datasets that capture co-speech gestures naturally aligned with surrounding objects, limiting progress toward gesture generation that is both communicative and physically grounded in real scenes.~\cite{nyatsanga2023review}

\section{Embody3D Preprocessing}
Embody3D~\cite{mclean2025embody} contains 500 hours of individual 3D human motion data from 439 participants. However, not all tasks are suitable for gesture generation. Among the categories, we select the Dyadic Conversation category, which contains 59.4 hours from 86 participants, and the Multi-person Conversation category, which contains 125.2 hours from 210 participants.

Using the selected subset of the Embody3D dataset introduces several challenges. (i) Participants occasionally transition between standing and sitting, creating unwanted motion segments. (ii) In some sequences, participants walk while speaking. (iii) In dyadic conversations there are two participants, and in multi-person conversations there are more than two; consequently, many frames correspond to participants listening rather than speaking, which are less useful for our task. (iv) In multi-speaker conversations, speech segments may contain long pauses caused by interruptions from other participants. To address these issues, we preprocess the data as follows.

\vspace{5pt}
\noindent \textbf{Audio Transcription and Word-Level Alignment.} The first step extracts speech transcripts and precise word timings from the raw audio. The audio is transcribed using WhisperX~\cite{bain2022whisperx}, which produces both sentence-level transcription and word-level timestamps. Word-level alignment allows accurate synchronization between speech and motion.  This alignment is later used to identify continuous speaking segments and associate them with motion frames.

\vspace{5pt}
\noindent \textbf{Speech Segment Extraction and Audio Segmentation.} Speech segments are extracted from the aligned word timestamps by grouping consecutive words into continuous speaking intervals. A new segment is created when the temporal gap between consecutive words exceeds a predefined threshold, and segments shorter than a minimum duration are discarded. For each valid segment, the corresponding portion of the audio waveform is trimmed using the start and end timestamps. This produces a set of audio clips representing continuous speaking intervals. The waveform sample indices are also preserved to enable alignment between the audio segments and the motion frames.

\vspace{5pt}
\noindent \textbf{Corrupted Motion Filtering.} Segments containing corrupted motion data are removed. The dataset provides a \texttt{missing\_frames} indicator specifying frames where motion capture failed. The audio sample indices are mapped to motion frame indices, and any segment containing missing frames is discarded to ensure that only complete motion sequences are retained. 

\vspace{5pt}
\noindent \textbf{Sitting and Standing Detection.} The body pose parameters are analyzed to determine whether the participant is sitting or standing. This is computed using the relative angles between the spine and hip joints. Segments that contain both sitting and standing frames are discarded to maintain consistent body posture within each sequence.

\vspace{5pt}
\noindent \textbf{Motion Canonicalization and Coordinate Conversion.} Motion data are transformed into a consistent coordinate system. The original Y-up coordinate system is converted to a Z-up coordinate system via a rotation around the x-axis. In addition, the global orientation is normalized to ensure that the actor faces a consistent direction across all sequences.

\vspace{5pt}
\noindent \textbf{Lower Body Stabilization for Standing Sequences.} For standing segments, lower body motion is stabilized to remove locomotion artifacts. The SMPL-X body model is used to reconstruct the skeleton, after which the lower body joints are reset to their rest pose using inverse kinematics~\cite{li2025hybrik}. This suppresses walking or stepping motions while preserving upper-body gestures.

\vspace{5pt}
\noindent \textbf{Motion Reconstruction and Data Export.} Finally, the processed motion sequences are reconstructed using the SMPL-X representation. For each valid segment, we export three synchronized modalities: the audio clip, the motion sequence, and the corresponding text annotations. The motion is stored as SMPL-X pose parameters together with the body shape coefficients and frame rate information. The text annotations contain the transcribed words along with their start and end timestamps relative to the segment, enabling precise alignment between speech and motion.

The resulting dataset contains \textbf{18,380 motion sequences}, covering approximately \textbf{43 hours of motion} from \textbf{280 unique speakers}. Each sample therefore consists of temporally aligned audio, text, and motion data, making it suitable for speech-driven gesture generation tasks. The processed dataset will be publicly released to facilitate future research on speech-driven human motion generation.

\section{More Details about Evaluation Metrics}
\subsection{FGD.} We use the same Fréchet Gesture Distance (FGD)~\cite{yoon2020speech} as in prior work. However, since our method neutralizes the lower body (\cref{app:canonicalization}), we ensure a fair comparison with previous approaches,which may produce occasional lower-body motion, by fixing the lower-body joints of the sequences to the identity pose rotation (i.e., no rotation) during evaluation. This effectively restricts the FGD comparison to upper-body motion. In contrast, BC~\cite{li2021ai} is already computed using upper-body joint locations only; therefore, no modification is required.
\subsection{Gesture Diversity.} To quantify gesture diversity for a given audio input, we generate $K=5$ motion samples and measure their variation. Let $\mathbf{X}\in\mathbb{R}^{K\times T\times J\times 3}$ denote the predicted joint positions, where $T$ is the number of frames and $J$ the number of joints. During evaluation, we restrict the analysis to upper-body joints $J_u$ and apply temporal smoothing using a moving-average filter with window size $w=5$. For each pair of generated samples $i$ and $j$, we compute the framewise L1 distance
\begin{equation}
D_{i,j}(t,j,c)=\left|X_{i,t,j,c}-X_{j,t,j,c}\right|,
\end{equation}
where $t$ indexes time, $j$ joints, and $c\in\{x,y,z\}$ coordinates. The mean pairwise distance across all $\frac{K(K-1)}{2}$ sample pairs is
\begin{equation}
D(t,j,c)=
\frac{2}{K(K-1)}
\sum_{i=1}^{K-1}\sum_{j=i+1}^{K} D_{i,j}(t,j,c),
\end{equation}
yielding $D\in\mathbb{R}^{T\times J_u\times 3}$.

To normalize the magnitude across joints and coordinates, we compute a per-dimension scale using the temporal median
\begin{equation}
s_{j,c}=\mathrm{median}_{t}\big(D(t,j,c)\big).
\end{equation}
The normalized diversity tensor is then
\begin{equation}
\tilde{D}(t,j,c)=\frac{D(t,j,c)}{s_{j,c}+\epsilon},
\end{equation}
where $\epsilon$ is a small constant for numerical stability. Finally, the gesture diversity score for the audio clip is obtained by averaging across time, joints, and coordinates
\begin{equation}
\mathcal{D}=\frac{1}{T J_u 3}\sum_{t=1}^{T}\sum_{j=1}^{J_u}\sum_{c=1}^{3}\tilde{D}(t,j,c).
\end{equation}

\section{Dual Facets of Object Awareness and Limitations.} 
\revise{We explicitly clarify that environment-grounded gesture generation encompasses two distinct aspects: (i) \textit{semantic and spatial correlation}, where the thematic style and orientation of conversational gestures are naturally influenced by surrounding furniture assets, and (ii) \textit{physical plausibility}, where synthesized motions strictly avoid environmental penetration. While \puppet{} models both facets simultaneously, achieving absolute, zero-collision physical plausibility and prefect object contact is still a formidable open challenge across the broader human-scene interaction (HSI) field~\cite{xing2026interphys, liu2026open, cai2025interactmove}. Consequently, \emph{minor} physical penetrations may still occur in a small fraction of frames. Nonetheless, the synthesized gestures remain visibly and structurally shaped by the surrounding objects. 

To the best of our knowledge, this is the first work to formulate co-speech gesture generation as a joint posture-aware and object-conditioned problem. As a first step, we make a simplifying design choice by focusing on stationary conversational gestures and removing root trajectory, rather than modeling gestures during walking or large body movement. As research in this emerging problem space advances and more suitable datasets become available, we expect these assumptions to be relaxed.
}




\begin{figure*}[t]
\centering
\begin{minipage}{1.0\linewidth}
  \begin{promptbox}{Scenario Creation Prompt}
    \vspace{5pt}
    \textbf{\normalsize Instructions.}\\[5pt]
    {
    \sffamily \scriptsize
    Generate \texttt{\textcolor{blue}{<|count|>}} conversational scenarios for two characters.\\[3pt]
        \textbf{\normalsize Rules:}\\ [2pt]
        - Realistic situations. \\
        - Make them diverse across domains, intents and scenes and scenarios.\\
        - Avoid repeating similar everyday small-talk situations. Be creative, explore different contexts and unique environments.\\
        - No two scenarios in the same batch should feel alike; each should introduce a distinct setting and dynamic.\\
        - Neither char description may rely on manipulating/using an object (e.g., “checking the schedule”, “assembling furniture”).\\
        - No scenario involving a child. All scenarios must involve adults.\\
        - The scenario must be meaningful as dialogue.\\
        - Scenarios should NOT involve object(s); removing all object dependency must still leave a clear social exchange.\\
        - Each scenario must have enough substance to sustain a multi-turn conversation (a few minutes of dialogue). Avoid trivial one-line exchanges (e.g., a quick joke and a short reply). \\
        - MUST include exactly one pose from one of: \texttt{sit\_no\_table}, \texttt{sit\_behind\_table}, \texttt{stand}, \texttt{stand\_behind\_table}. You SHOULD generate based on this number of poses \texttt{\textcolor{blue}{<|Goal\_pose\_dist|>}}. \\
        
        \textbf{\normalsize Json rules:}\\
        - Short fields only.\\
        - Each must include a *scene* (<= 12 words).\\
        - Characters with short role descriptions (<= 12 words each).\\
        - Assign gender for each (male/female only).\\
        - One 1–2 word 'scenevibe' describing the atmosphere.\\
        - One pose from: \texttt{sit\_no\_table}, \texttt{sit\_behind\_table}, \texttt{stand}, \texttt{stand\_behind\_table}.\\
        Return JSON ONLY that conforms to this schema:
        {\ttfamily
        \scriptsize
        \begin{verbatim}
        {
          "items": [
            {
              "scene": "string <= 12 words, specific but brief e.g. interrogation room",
              "char1": "string <= 12 words, e.g. detective interrogating a suspect",
              "char2": "string <= 12 words, e.g. suspect being interrogated",
              "char1_gender": "male OR female",
              "char2_gender": "male OR female",
              "pose": "sit_no_table OR sit_behind_table OR stand OR stand_behind_table",
              "scenevibe": "string, 1–2 words, (e.g., calm, tense, warm, neutral, urgent, 
              playful, professional, friendly, formal, informal)"
            }
          ]
        }
        \end{verbatim}
        }
        }
    \textbf{\large System Promopt.}\\[8pt]
    \sffamily \scriptsize
        You generate conversational scenarios for multi-speaker conversational datasets. \\
        Output MUST be valid JSON only (no prose).  \\
        Each item must describe a setting (scene) and two interacting roles: char1 and char2. \\ 
        Only conversational scenarios. No solitary scenes.\\
        All items must imply dialogue/interaction; exclude solitary or purely reflective situations. \\
        Scenarios must stand alone without objects involved; removing all object dependency must still leave a clear social exchange.\\
        Ban object activities (e.g., assembling, fixing, checking, carrying, unlocking, typing, scanning, paying, cooking)\\
        Scenarios must be non-sensitive, broadly safe for all ages. \\
        Avoid medical diagnoses, legal advice, financial advice, politics, adult content, or PII. \\
        No scenario involving a child. All scenarios must involve adults.\\
  \end{promptbox}
\end{minipage}\vspace{-5pt}
\caption{\scriptsize System prompt for scenario creation for \sceneges. A scenario defines a scene (setting), two interacting roles, a pose constraint (e.g., sit, stand), and a scene atmosphere, returned in structured JSON format.}
\label{fig:scenario_prompt}
\end{figure*}

\begin{figure*}[t]
\centering
\begin{minipage}{1.0\linewidth}
  \begin{promptbox}{Dialogue Creation Prompt}
    \vspace{-2pt}
\textbf{\normalsize Instructions.}\\[5pt]
{
    {
    \sffamily \scriptsize
    Write a naturalistic two-character dialogue for this conversational setting. This script is going to be used for video generation for the task of gesture generation from audio. So try to use descriptive language that can convey gestures.\\[1pt]

    Scene: \texttt{\textcolor{blue}{<|scene|>}}\\
    Character 1 (\texttt{\textcolor{blue}{<|char1\_gender|>}}): \texttt{\textcolor{blue}{<|char1|>}}\\
    Character 2 (\texttt{\textcolor{blue}{<|char2\_gender|>}}): \texttt{\textcolor{blue}{<|char2|>}}\\
    Pose for char1 and char2: \texttt{\textcolor{blue}{<|pose|>}}\\
    General scene vibe: \texttt{\textcolor{blue}{<|scenevibe|>}}\\[3pt]
    
    \textbf{\normalsize Hard constraints:}\\[2pt]
    - Dialogue only. No narration/stage directions/SFX.\\
    - Keep the whole script $\leq$ \texttt{\textcolor{blue}{<|all\_max\_words|>}} words total.\\
    - Alternate turns between speakers.\\
    - Do not invent new characters; only char1 and char2.\\
    - IMPORTANT: Split the dialogue into contiguous chunks; each chunk must have between \texttt{\textcolor{blue}{<|max\_words\_min|>}} and \texttt{\textcolor{blue}{<|max\_words|>}} words.\\
    - At the START of every line: include the speaker tag \texttt{(char1)} or \texttt{(char2)}, the tone \texttt{(TONE:<tone>)}, and the emotion \texttt{(EMOTION:<emotion>)}.\\
    - Emotion should be one of: \texttt{neutral}, \texttt{anger}, \texttt{happiness}, \texttt{fear}, \texttt{disgust}, \texttt{sadness}, \texttt{surprise}, \texttt{contempt}.\\
    - Bind to the following format: \\
    {\ttfamily
    \scriptsize
    \begin{verbatim}
    <FULL SCRIPT>
    (EMOTION:<emotion>) (TONE:<tone>) (char1) ...
    (EMOTION:<emotion>) (TONE:<tone>) (char2) ...
    (EMOTION:<emotion>) (TONE:<tone>) (char1) ...
    ...
    [FINISHED]
    
    chunk1: (EMOTION:<emotion>) (TONE:<tone>) (char<NUMBER>) <dialogue lines for chunk 1>
    chunk2: (EMOTION:<emotion>) (TONE:<tone>) (char<NUMBER>) <dialogue lines for chunk 2>
    chunk3: ...
    \end{verbatim}
    }
    - Example format: \\
    {\ttfamily
    \scriptsize
    \begin{verbatim}
    <FULL SCRIPT>
    (EMOTION:<neutral>) (TONE:academic) (char1) <DIALOGUE>
    (EMOTION:<neutral>) (TONE:neutral) (char2) <DIALOGUE>
    (EMOTION:<neutral>) (TONE:neutral) (char1) <DIALOGUE>
    (EMOTION:<surprise>) (TONE:Inquisitive) (char2) <DIALOGUE>
    [FINISHED]
    
    chunk1: (EMOTION:<neutral>) (TONE:academic) (char1) <DIALOGUE>
    chunk2: (EMOTION:<neutral>) (TONE:neutral) (char2) <DIALOGUE>
    chunk3: (EMOTION:<neutral>) (TONE:neutral) (char1) <DIALOGUE>
    chunk4: (EMOTION:<surprise>) (TONE:Inquisitive) (char2) <DIALOGUE>
    \end{verbatim}
    }
    }
}
    \textbf{\normalsize System Promopt.}\\[8pt]
    \sffamily \scriptsize
        You are an award-winning screenwriter. \\
        Always obey hard limits. Output PLAIN DIALOGUE only—no narration, \\
        no stage directions, no actions, no SFX. \\
        Every utterance must start with a speaker tag: (char1) or (char2). \\
        Mark tone changes exactly with (TONE:\texttt{<tone>}). \\
        If constraints conflict, prioritize format correctness.\\
  \end{promptbox}
\end{minipage}\vspace{-10pt}
\caption{\scriptsize System prompt to generate multi-turn dialogue scripts for each scenario for \sceneges. The prompt conditions on the scene, roles, pose, and atmosphere, and enforces structured dialogue with speaker tags, tone, emotion labels, and chunked segments for video generation.}
\label{fig:dialogue_prompt}
\end{figure*}

\begin{figure*}[t]
\centering
\begin{minipage}{1.0\linewidth}
  \begin{promptbox}{Granular Gesture Suggestion Prompt}
    \vspace{5pt}
\textbf{\normalsize Instructions.}\\[5pt]
{
    {
        \sffamily \scriptsize
        Role as an expert prompt engineer for Google's Veo model that wants to generate video of a character saying a dialogue.\\[3pt]
        
        You are given a text. Scene is \texttt{\textcolor{blue}{<|scene|>}} and scene vibe is \texttt{\textcolor{blue}{<|scenevibe|>}}.\\
        
        For this text, suggest gestures for the motions of a \texttt{\textcolor{blue}{<|speaker\_gender|>}} character with respect to its context state. Gesture should be realistic. Do not exaggerate gestures.\\
        
        Suggest gestures based on keywords triggering gestures and semantics for someone in pose \texttt{\textcolor{blue}{<|pose|>}}.\\
        
        The character is saying the text to another character with emotion \texttt{\textcolor{blue}{<|emotion|>}}. Wherever needed, insert gesture suggestions inside \texttt{``[ ]''} within the text. Based on this text, a video model will generate the video of the character.\\[3pt]
        
        \textbf{\normalsize Gesture suggestion rules:}\\ \vspace{5pt}
        
        - Be affirmative (do not use words like ``perhaps'', ``might'', etc.).\\
        - Do not use "or".\\
        - Do not refer to any objects beyond what is mentioned in the context state.\\[3pt]
        
        Input text: \texttt{\textcolor{blue}{<|chunk\_text|>}}
        }
    }
  \end{promptbox}
\end{minipage}
\caption{\scriptsize System prompt to generate granular gesture suggestions for each dialogue chunk. The prompt conditions on the scene, pose, emotion, and speaker context, and inserts realistic object-aware gesture descriptions directly into the dialogue text for video generation.}
\label{fig:gesture_prompt}
\end{figure*}

\begin{figure*}[t]
\centering
\begin{minipage}{1.0\linewidth}
  \begin{promptbox}{Video Generation Prompt}
    \vspace{5pt}
\textbf{\normalsize Instructions.}\\[5pt]
{
    \sffamily \scriptsize
        (VERY IMPORTANT: Static shot, fixed camera) The character maintains the same position and stance throughout the entire take. The person shows clear body gestures with respect to the dialogue and looks directly into the camera, talking to a hypothetical person.\\
        
        The character says with emotion \texttt{\textcolor{blue}{<|emotion|>}} the dialogue: \texttt{\textcolor{blue}{<|dialogue|>}}. These inputs are produced from the previous generation steps in the pipeline.\\[3pt]
        
        \textbf{\normalsize Gesture rules:}\\
        
        - Bracketed cues in the dialogue mark gesture suggestions. Generate gestures based on these cues.\\
        - Body gestures appear only during the dialogue phrase.\\[3pt]
        
        \textbf{\normalsize Hard constraints:}\\
        
        - Use only the provided image as the scene, unchanged.\\
        - The image background, set of objects, and scene remain consistent throughout. Any pointing or referring is hypothetical.\\
        - VERY IMPORTANT: The character stays in place without walking or moving around.\\
        - Hands should always remain within the frame and must not be cropped.\\
        - The take is continuous with steady lighting and without visual overlays.\\
        - Audio track includes only the spoken dialogue.\\
        - Align mouth movement exactly with the dialogue text.\\
        - The visual remains free of text and graphic overlays.\\[3pt]
        
        \textbf{\normalsize Output intent:}\\
        - A single continuous take that looks like a person speaking directly to the camera (static shot, fixed camera).

    }
  \end{promptbox}
\end{minipage}
\caption{\scriptsize System prompt used for video generation. It conditions the video model on the dialogue, emotion, and gesture cues, while enforcing a static camera, fixed scene layout, and gesture execution aligned with the dialogue.}
\label{fig:video_prompt}
\end{figure*}

\end{document}